\documentclass{article}

\usepackage[final,main]{neurips_2026}

\usepackage[utf8]{inputenc} 
\usepackage[T1]{fontenc}    
\usepackage{hyperref}       
\usepackage{url}            
\usepackage{booktabs}       
\usepackage{amsfonts}       
\usepackage{nicefrac}       
\usepackage[nopatch=footnote]{microtype}      
\usepackage[table]{xcolor}  
\usepackage{fontawesome5}

\usepackage{amsmath}
\usepackage{amssymb}
\usepackage{amsthm}
\usepackage{mathtools}
\usepackage{bm}

\usepackage{graphicx}
\usepackage{caption}
\usepackage{subcaption}
\usepackage{multirow}
\usepackage{array}
\usepackage{tabularx}
\usepackage{threeparttable}
\usepackage{makecell}
\usepackage{siunitx}
\usepackage{float}
\usepackage{wrapfig}

\usepackage{algorithm}
\usepackage{algpseudocode}

\usepackage{enumitem}
\usepackage{setspace}

\usepackage[nameinlink]{cleveref}

\DeclareCaptionLabelSeparator{pipe}{\space|\space}
\newtheorem{proposition}{Proposition}

\newtheorem{corollary}{Corollary}
\theoremstyle{definition}

\theoremstyle{remark}

\title{Lifted State Policy Optimization: Internalizing Self-Correction through Transition Geometry}

\author{
  Ren Zhuang \quad
  Ben Wang\textsuperscript{\dag} \quad
  Shuifa Sun\textsuperscript{\dag} \vspace{1mm} \\
  Hangzhou Normal University \vspace{1mm} \\
  \texttt{venomrko97@gmail.com, \{20170056, watersun\}@hznu.edu.cn}
  \vspace{-2mm}
}

\usepackage[most]{tcolorbox}
\begin{document}

\begingroup
\renewcommand\thefootnote{\dag} 
\footnotetext{Corresponding author.}
\endgroup

\maketitle

\begin{abstract}
Preference optimization ranks final answers and trajectory RL maximizes aggregate returns, yet neither credits the local change that turns a flawed reasoning state into a better one. We introduce \textit{Lifted State Policy Optimization (LSPO)}, which establishes this finite, cost-adjusted transition as the primary unit of learning. Each textual answer is augmented with a learned continuous coordinate, forming a lifted state that an energy function evaluates and orders. A shaped reward drives energy descent net of edit and step costs, and an internalization loss absorbs selected low-energy answers into the direct generator. On Qwen3.5-27B, the resulting single-pass policy reaches 71.8\% aggregate accuracy across mathematics, science, code, and logic benchmarks, matching test-time reflection at 71.9\% with 28\% lower latency and 25.8\% fewer tokens. The pattern holds across dense and mixture-of-experts backbones, and matched controls confirm that transition credit and the lifted coordinate each contribute, as does answer internalization. Our code is available at \url{https://github.com/Venomrko/lspo}.
\end{abstract}

\begin{figure}[H]
    \centering
    \captionsetup{font=small}
    \includegraphics[width=0.98\textwidth]{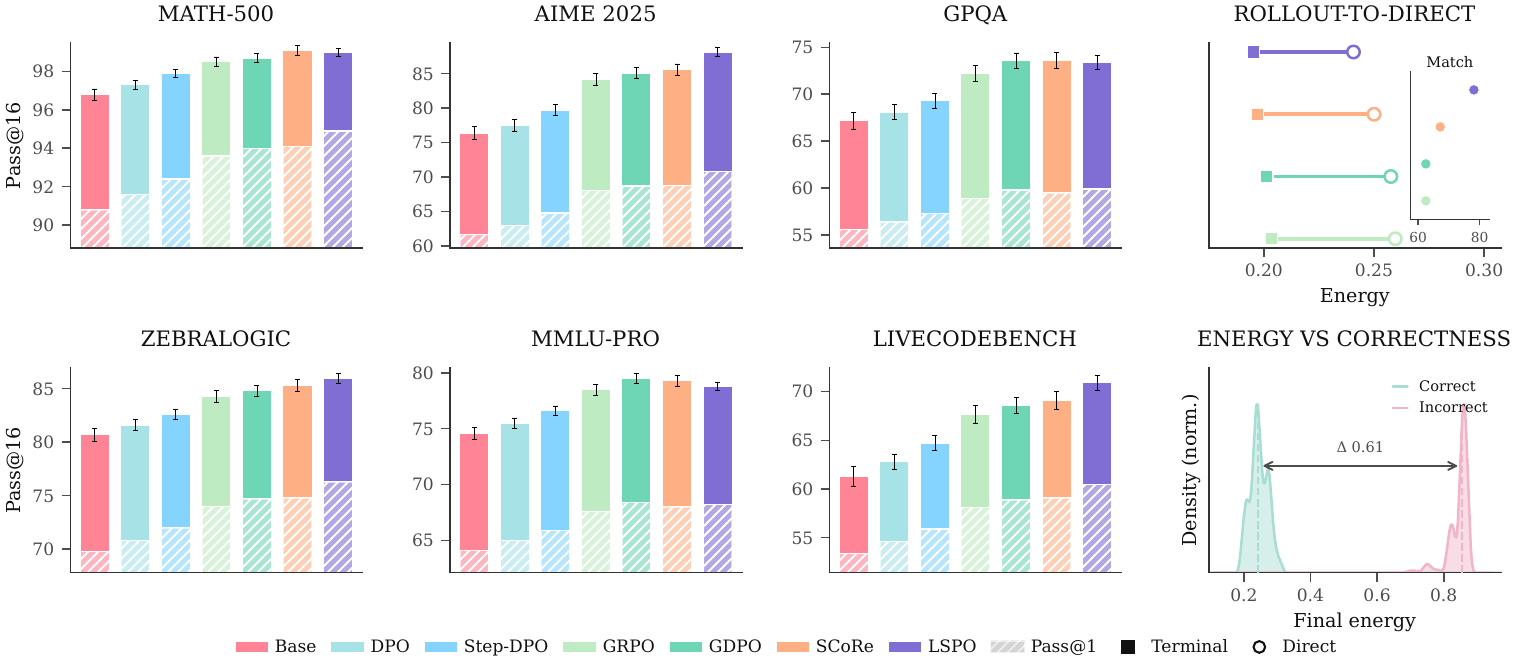}
\caption{Improvements found during training persist in direct generation. On Qwen3.5-27B, the LSPO policy attains the highest aggregate direct pass@1 and pass@16 accuracy across benchmarks in the left panels, selected training endpoints align closely with deployed outputs in the top right, and final energy separates correct from incorrect answers in the bottom right.}
    \label{fig:teaser_overview}
\end{figure}

\section{Introduction}

Even strong language models make local mistakes within otherwise sound reasoning, and one faulty step can invalidate the final answer \citep{uesato2022solving, lightman2023let}. Inference methods address this through chain of thought, self-consistency, critique, and repeated refinement \citep{wei2022chain, wang2022self, madaan2023self, shinn2023reflexion}, and large reasoning models extend deliberation into long trajectories \citep{shao2024deepseekmath, zelikman2024quiet, guo2025deepseek}. These procedures improve difficult answers, but their gains require additional sampling budgets or persistent refinement loops at deployment. Training can absorb useful corrections into the policy itself, removing this cost.

A central question is defining the appropriate unit of supervision. Current objectives provide supervision at coarse granularity; preference optimization ranks complete outputs \citep{rafailov2023direct, hong2024orpo, meng2024simpo, liu2025survey}, trajectory RL maximizes aggregate returns over full rollouts \citep{ouyang2022training, schulman2017proximal}, and self-correction methods learn from entire refinement traces \citep{kumar2024training, ma2025s2r}. Each focuses on endpoints or full trajectories, leaving the local transition that resolves a flawed reasoning state as an unmodeled unit of credit. We formalize this gap through \emph{transition credit}, where the unit of learning is a finite state transition, evaluated by the decrease in a learned energy function net of edit and step costs \citep{sutton1998reinforcement, kaufmann2024survey, lambert2025reinforcement}.

We introduce \textbf{L}ifted \textbf{S}tate \textbf{P}olicy \textbf{O}ptimization (\textbf{LSPO}). LSPO augments the textual answer $x$ with a continuous coordinate $z$, forming a lifted state $s=(x,z)$ that carries refinement context beyond discrete surface text. An energy function $\tilde{E}(s;q)$ orders candidate states; a shaped reward favors transitions that decrease energy net of edit and step costs; and an internalization loss trains the generator to produce selected low-energy answers directly from the original input.

We evaluate LSPO across the Qwen3.5 family \citep{qwen3.5} on mathematics \citep{hendrycks2021measuring, AIME25}, science \citep{rein2024gpqa, wang2024mmlu}, executable code \citep{jain2024livecodebench}, and logic \citep{lin2025zebralogic}. On Qwen3.5-27B, the single-pass policy reaches 71.8\% aggregate accuracy, matching test-time reflection at 71.9\% with 28\% lower latency and 25.8\% fewer tokens. The pattern holds across scales, reaching 60.8\% at 9B and 72.6\% on the 35B-A3B mixture-of-experts backbone. At the mechanism level, as reported in \Cref{fig:teaser_overview}, LSPO converts 6.4\% of incorrect answers to correct while regressing only 1.7\%, yielding a 4.7-point net transition gain.

Our core contributions are threefold.
\begin{itemize}
    \item \textbf{Single-pass reliability.} Absorbing selected low-energy endpoints relocates deliberation from inference into training; the deployed policy reaches reflective accuracy in a single pass, across dense and mixture-of-experts backbones.
    \item \textbf{Cost-adjusted transition credit.} LSPO establishes a finite state transition as the primary unit of learning, scored by energy decrease net of edit and step costs. Unlike a step-level score, this quantity is relative to a prompt's candidate pool, steering policy movement rather than global state ranking. Bounded energy error perturbs cumulative credit only through the endpoints, and removing the progress reward costs more net gain than removing any other component.
    \item \textbf{Lifting as refinement context.} A learned continuous coordinate carries information about constructive transitions that surface text alone does not; probes on held-out transitions show higher action prediction and energy ranking, while shuffled or zeroed coordinates eliminate this advantage.
\end{itemize}

\section{Method}
\label{sec:method}

\subsection{Overview}

\begin{figure*}[t]
    \centering
    \includegraphics[width=0.98\textwidth]{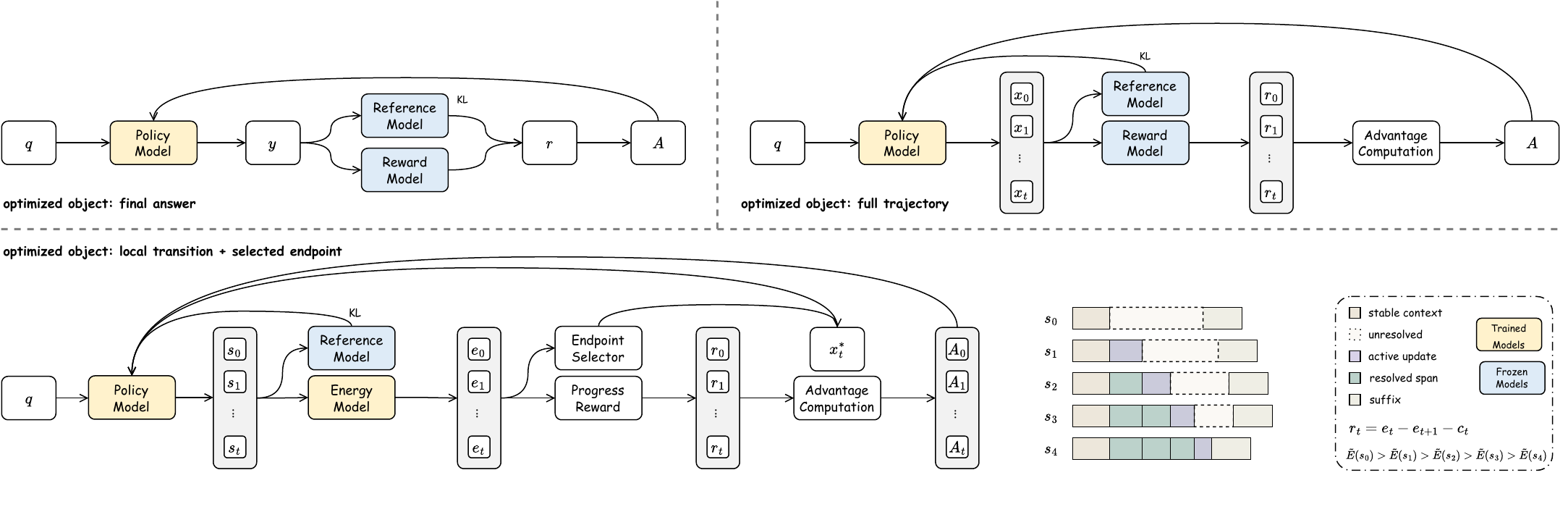}
    \caption{Comparison of training paradigms. Preference learning in the top left panel optimizes static final answers, while trajectory RL in the top right panel maximizes returns across full rollouts. LSPO in the bottom panel assigns credit to local transitions and internalizes selected endpoints.}
    \label{fig:method_compare}
\end{figure*}

LSPO learns local directions of improvement in a lifted state space and transfers selected answers into a direct policy. It combines a continuous coordinate for refinement context, an energy objective for crediting finite state changes, and an internalization loss that trains the generator on selected answers. \Cref{fig:method_compare} contrasts this localized mechanism with paradigms centered on terminal answers or aggregate returns.

\subsection{Lifted State}

Standard optimization procedures target a distribution $p_\theta(y \mid q)$ over complete outputs for a given input $q$, masking the intermediate dynamics of local revisions. To expose these dynamics for training, we cast local improvement as a Markov decision process in an augmented state space. We define a lifted state
\begin{equation}
    s_t = (x_t, z_t),
\end{equation}
where $x_t$ represents the surface text and $z_t$ is a continuous auxiliary coordinate. The auxiliary coordinate is initialized from the input and current answer, updated by the transition block, and projected away when the final answer is written back into text.

We introduce an energy function
\begin{equation}
    \tilde{E}(s_t; q) = \tilde{E}(x_t, z_t; q),
\end{equation}
to evaluate state quality. The energy head orders candidate states and defines which answer becomes the target for direct generation. The policy induces a transition distribution $P_\theta(s_{t+1} \mid s_t, a_t, q)$. Training optimizes the action policy $\pi_\theta(a_t \mid s_t, q)$ to favor short sequences whose transitions reduce energy.

Because surface text is discrete, moves are finite and are evaluated within visited neighborhoods.

\subsection{Training Procedure}

The training phase constructs short moves in the lifted state space. At each step, the policy samples an action $a_t \in \{\text{revise}, \text{switch}, \text{stop}\}$ from $\pi_{\theta}(a_t \mid s_t, q)$ to modify the surface text, move the auxiliary coordinate toward an alternative refinement mode, or halt within the current local basin. The frozen projection initializes $z_0$; Revise recursively updates $z_t$, and Switch resets $z_t$ before a residual update. The primary configuration uses $d_z=8$, and the energy head scores the resulting state. The resulting sequence is $\tau = (s_0, a_0, s_1, a_1, \dots, s_T)$. \Cref{app:ablation_stability} defines the edit metric, rollout collection, advantage estimator, and complete training procedure.

\subsection{Energy Model and Selection}

The energy model establishes an ordering over candidate states that guides both local transitions and final answer selection. It operates as a scalar head over $(q,x_t,z_t)$, shared across training families and fitted on sampled states. Gold answers and executable checkers provide scalar supervision. Verified candidates receive lower energy than unsuccessful ones, with ties broken by shorter edit distance and lower step cost. Pairs are formed within each prompt's candidate rollouts, assigning lower energy to verified states than to failed or perturbed alternatives.

Selection uses the same object. The generated sequence induces a candidate set $\mathcal C(q)$ containing the direct answer state and its successors. The method selects the state with the lowest energy
\begin{equation}
    s^*(q)
    =
    \arg\min_{s\in\mathcal C(q)} \tilde E(s;q)
    \label{eq:energy_selection}
\end{equation}
 and sets $x_T^*(q)$ to be its surface text component as the transfer target. The model is trained to emit $x_T^*(q)$ from the original input. Before policy training, we fit and freeze the entire scoring branch, encompassing the text encoder, lift projection, transition block, and energy head, as detailed in \Cref{app:ablation_stability}.

\subsection{Objective}

The LSPO objective unifies shaped reinforcement learning over intermediate transitions with answer internalization. The reward for stepwise progress is formulated as
\begin{equation}
    r_t =
    \tilde E(s_t; q) - \tilde E(s_{t+1}; q)
    - \lambda_{\mathrm{edit}} d(x_{t+1}, x_t)
    - \lambda_{\mathrm{step}},
    \label{eq:progress_reward}
\end{equation}
for Revise and Switch. Stop preserves the current state and yields zero reward. Here $d$ penalizes surface edits and $\lambda_{\mathrm{step}}$ regularizes trajectory length. We index the $T$ text-changing moves by $t=0,\ldots,T-1$; an optional final Stop contributes zero to their sum.

\begin{proposition}[Telescoping Trajectory Reward]
\label{prop:reward_decomposition}
For an input $q$ and a finite trajectory $\tau = (s_0, a_0, \dots, s_T)$, the cumulative shaped reward $R(\tau; q) = \sum_{t=0}^{T-1} r_t$ satisfies
\begin{equation}
    R(\tau; q)
    =
    \tilde E(s_0; q) - \tilde E(s_T; q)
    - \lambda_{\mathrm{edit}} \sum_{t=0}^{T-1} d(x_{t+1}, x_t)
    - \lambda_{\mathrm{step}} T.
    \label{eq:telescoping_reward}
\end{equation}
\end{proposition}
Because intermediate energy terms telescope, cumulative return reflects net endpoint improvement minus total accrued penalties. The proof appears in \Cref{app:ablation_stability}.

The same telescoping structure provides robustness to scoring error. When training energy differs from $\tilde E$ by at most $\varepsilon$ at visited states, cumulative reward shifts by at most $2\varepsilon$ through the endpoints alone, preserving the ranking of trajectories separated by more than $2\varepsilon$, as established in \Cref{prop:noisy_energy_robustness} and \Cref{cor:noisy_energy_ranking}.

At update $k$, the rollout snapshot $\bar\theta_k$ fixes the initial-state distribution $\rho_{\bar\theta_k}$ and text-proposal kernel $K_{\bar\theta_k}$. Controller RL uses the per-update objective
\begin{equation}
    J_{\mathrm{ctrl}}^{(k)}(\theta)
    = \mathbb{E}_{\tau\sim P_{\rho_{\bar\theta_k},\pi_\theta,K_{\bar\theta_k}}}
    \Big[\sum_t r_t-\beta\sum_t D_{\mathrm{KL}}\!\left(\pi_\theta(\cdot\mid s_t,q)\|\pi_{\mathrm{ref}}(\cdot\mid s_t,q)\right)\Big],
    \label{eq:rl_objective}
\end{equation}
where the action KL includes Stop. The initialization and proposal mechanisms are held fixed within an update and refreshed for the next one. The visited-state KL follows regularized policy optimization \citep{schulman2015high, schulman2017proximal, ouyang2022training, guo2025deepseek}.

To absorb the selected improvements into the generator, we incorporate a loss for selected internalization
\begin{equation}
    J_{\mathrm{SI}}(\theta)
    =
    \mathbb{E}_{q}
    \left[
        \log p_\theta(x_T^*(q) \mid q)
    \right],
    \label{eq:si_objective}
\end{equation}
where $x_T^*(q)$ denotes the selected final output with the lowest energy produced during the alignment phase. The final objective is
\begin{equation}
    J_{\mathrm{LSPO}}^{(k)}(\theta)
    =
    J_{\mathrm{ctrl}}^{(k)}(\theta)
    +
    \alpha J_{\mathrm{SI}}(\theta).
    \label{eq:lspo_objective}
\end{equation}

Task feedback supplies the supervision used to fit the energy ordering.

Formal properties of the objective appear in the appendices. In particular, the internalization loss corresponds to a forward KL projection onto the selected answer distribution in \Cref{app:internalization}, while \Cref{app:theory_roadmap} gives a path-distribution relaxation and an information criterion for useful lifting.

\label{sec:deployment_interface}
At deployment, the compiled conditional distribution serves each answer through direct generation, requiring no auxiliary lifted-state computation during inference.

\section{Experiments}
\label{sec:experiments}

\newcommand{\modeltag}[2]{%
    \multirow{#1}{*}{\rotatebox{90}{#2}}%
}
\begin{table*}[t]
    \centering
    \caption{Main results across Qwen3.5 backbones. LSPO achieves the highest aggregate accuracy among the listed baselines that generate directly under both pass@1 and pass@16 evaluation, with frontier models included as scale references.}
    \label{tab:main_results_all_models}
    \footnotesize
    \resizebox{\textwidth}{!}{%
    \begin{tabular}{l|lrrrrrrrrrrrrrr}
        \toprule
        \textbf{Model} & \textbf{Method} & \multicolumn{2}{c}{\textbf{MATH-500}} & \multicolumn{2}{c}{\textbf{AIME25}} & \multicolumn{2}{c}{\textbf{GPQA}} & \multicolumn{2}{c}{\textbf{LCB}} & \multicolumn{2}{c}{\textbf{ZebraLogic}} & \multicolumn{2}{c}{\textbf{MMLU-Pro}} & \multicolumn{2}{c}{\textbf{Average}} \\
        \cmidrule(lr){3-4}\cmidrule(lr){5-6}\cmidrule(lr){7-8}\cmidrule(lr){9-10}\cmidrule(lr){11-12}\cmidrule(lr){13-14}\cmidrule(lr){15-16}
        & & \textbf{@1} & \textbf{@16} & \textbf{@1} & \textbf{@16} & \textbf{@1} & \textbf{@16} & \textbf{@1} & \textbf{@16} & \textbf{@1} & \textbf{@16} & \textbf{@1} & \textbf{@16} & \textbf{@1} & \textbf{@16} \\
        \midrule
        \modeltag{7}{Qwen3.5-9B} & Base & 84.8 & 93.6 & 39.6 & 57.2 & 45.8 & 61.4 & 41.9 & 48.6 & 60.2 & 73.4 & 56.3 & 68.8 & 54.8 & 67.2 \\
         & DPO & 85.6 & 94.1 & 40.9 & 58.3 & 46.6 & 62.3 & 43.1 & 50.1 & 61.2 & 74.3 & 57.2 & 69.7 & 55.8 & 68.1 \\
         & Step-DPO & 86.4 & 94.7 & 42.7 & 60.5 & 47.5 & 63.5 & 44.4 & 52.0 & 62.4 & 75.3 & 58.1 & 70.8 & 56.9 & 69.5 \\
         & GRPO & 87.6 & 95.3 & 46.0 & 65.0 & 49.1 & 66.4 & 46.6 & 55.0 & 64.4 & 77.0 & 59.8 & 72.7 & 58.9 & 71.9 \\
         & GDPO & \underline{88.0} & 95.5 & \underline{46.6} & 65.9 & \underline{50.0} & \textbf{67.8} & \underline{47.4} & 55.9 & \underline{65.1} & 77.5 & \textbf{60.6} & \textbf{73.7} & \underline{59.6} & 72.7 \\
         & SCoRe & \underline{88.0} & \underline{95.8} & 46.4 & \underline{66.1} & 49.5 & 67.6 & \underline{47.4} & \underline{56.2} & 65.0 & \underline{77.8} & 60.0 & \underline{73.3} & 59.4 & \underline{72.8} \\
         & \cellcolor[HTML]{EDE7F6}LSPO (\textbf{Ours}) & \cellcolor[HTML]{EDE7F6}\textbf{89.0} & \cellcolor[HTML]{EDE7F6}\textbf{95.9} & \cellcolor[HTML]{EDE7F6}\textbf{49.0} & \cellcolor[HTML]{EDE7F6}\textbf{69.2} & \cellcolor[HTML]{EDE7F6}\textbf{50.2} & \cellcolor[HTML]{EDE7F6}\underline{67.7} & \cellcolor[HTML]{EDE7F6}\textbf{49.2} & \cellcolor[HTML]{EDE7F6}\textbf{58.4} & \cellcolor[HTML]{EDE7F6}\textbf{66.9} & \cellcolor[HTML]{EDE7F6}\textbf{78.9} & \cellcolor[HTML]{EDE7F6}\underline{60.5} & \cellcolor[HTML]{EDE7F6}73.1 & \cellcolor[HTML]{EDE7F6}\textbf{60.8} & \cellcolor[HTML]{EDE7F6}\textbf{73.9} \\
        \midrule
        \modeltag{7}{Qwen3.5-27B} & Base & 90.8 & 96.8 & 61.7 & 76.4 & 55.6 & 67.2 & 53.4 & 61.3 & 69.8 & 80.7 & 64.1 & 74.6 & 65.9 & 76.2 \\
         & DPO & 91.6 & 97.3 & 63.0 & 77.5 & 56.4 & 68.1 & 54.6 & 62.8 & 70.8 & 81.6 & 65.0 & 75.5 & 66.9 & 77.1 \\
         & Step-DPO & 92.4 & 97.9 & 64.8 & 79.7 & 57.3 & 69.3 & 55.9 & 64.7 & 72.0 & 82.6 & 65.9 & 76.6 & 68.0 & 78.5 \\
         & GRPO & 93.6 & 98.5 & 68.1 & 84.2 & 58.9 & 72.2 & 58.1 & 67.7 & 74.0 & 84.3 & 67.6 & 78.5 & 70.0 & 80.9 \\
         & GDPO & 94.0 & 98.7 & 68.7 & 85.1 & \underline{59.8} & \textbf{73.6} & 58.9 & 68.6 & 74.7 & 84.8 & \textbf{68.4} & \textbf{79.5} & \underline{70.8} & 81.7 \\
         & SCoRe & \underline{94.1} & \textbf{99.1} & \underline{68.8} & \underline{85.6} & 59.5 & \textbf{73.6} & \underline{59.1} & \underline{69.1} & \underline{74.8} & \underline{85.3} & 68.0 & \underline{79.3} & 70.7 & \underline{82.0} \\
         & \cellcolor[HTML]{EDE7F6}LSPO (\textbf{Ours}) & \cellcolor[HTML]{EDE7F6}\textbf{94.9} & \cellcolor[HTML]{EDE7F6}\underline{99.0} & \cellcolor[HTML]{EDE7F6}\textbf{70.8} & \cellcolor[HTML]{EDE7F6}\textbf{88.1} & \cellcolor[HTML]{EDE7F6}\textbf{59.9} & \cellcolor[HTML]{EDE7F6}\underline{73.4} & \cellcolor[HTML]{EDE7F6}\textbf{60.5} & \cellcolor[HTML]{EDE7F6}\textbf{70.9} & \cellcolor[HTML]{EDE7F6}\textbf{76.3} & \cellcolor[HTML]{EDE7F6}\textbf{86.0} & \cellcolor[HTML]{EDE7F6}\underline{68.2} & \cellcolor[HTML]{EDE7F6}78.8 & \cellcolor[HTML]{EDE7F6}\textbf{71.8} & \cellcolor[HTML]{EDE7F6}\textbf{82.7} \\
        \midrule
        \modeltag{7}{Qwen3.5-35B-A3B} & Base & 91.9 & 97.2 & 65.4 & 79.8 & 57.1 & 68.9 & 55.8 & 63.6 & 71.4 & 81.8 & 65.2 & 75.1 & 67.8 & 77.7 \\
         & DPO & 92.7 & 97.7 & 66.7 & 80.9 & 57.9 & 69.8 & 57.0 & 65.1 & 72.4 & 82.7 & 66.1 & 76.0 & 68.8 & 78.7 \\
         & Step-DPO & 93.5 & 98.3 & 68.5 & 83.1 & 58.8 & 71.0 & 58.3 & 67.0 & 73.6 & 83.7 & 67.0 & 77.1 & 70.0 & 80.0 \\
         & GRPO & 94.4 & 98.7 & 71.2 & 86.9 & 60.1 & 73.5 & 60.0 & 69.4 & 75.2 & 85.0 & 68.4 & 78.7 & 71.6 & 82.0 \\
         & GDPO & \underline{94.8} & \underline{98.9} & \underline{71.8} & 87.8 & \textbf{61.0} & \textbf{74.9} & \underline{60.9} & 70.4 & \underline{75.9} & 85.5 & \textbf{69.2} & \textbf{79.7} & \underline{72.3} & \underline{82.9} \\
         & SCoRe & 94.7 & \textbf{99.0} & 71.5 & \underline{87.9} & 60.5 & \underline{74.7} & 60.7 & \underline{70.5} & 75.7 & \underline{85.7} & \underline{68.6} & \underline{79.3} & 71.9 & 82.9 \\
         & \cellcolor[HTML]{EDE7F6}LSPO (\textbf{Ours}) & \cellcolor[HTML]{EDE7F6}\textbf{95.2} & \cellcolor[HTML]{EDE7F6}98.8 & \cellcolor[HTML]{EDE7F6}\textbf{72.9} & \cellcolor[HTML]{EDE7F6}\textbf{89.7} & \cellcolor[HTML]{EDE7F6}\underline{60.6} & \cellcolor[HTML]{EDE7F6}74.1 & \cellcolor[HTML]{EDE7F6}\textbf{61.6} & \cellcolor[HTML]{EDE7F6}\textbf{71.7} & \cellcolor[HTML]{EDE7F6}\textbf{76.8} & \cellcolor[HTML]{EDE7F6}\textbf{86.1} & \cellcolor[HTML]{EDE7F6}68.5 & \cellcolor[HTML]{EDE7F6}78.5 & \cellcolor[HTML]{EDE7F6}\textbf{72.6} & \cellcolor[HTML]{EDE7F6}\textbf{83.1} \\
        \midrule
        \multicolumn{2}{l}{gpt-oss-120b} & 96.3 & 99.0 & 81.8 & 91.6 & 75.6 & 82.4 & 61.7 & 70.9 & 80.4 & 87.5 & 80.8 & 86.1 & 79.4 & 86.2 \\
        \multicolumn{2}{l}{Claude Sonnet 4.5} & 97.0 & 99.1 & 87.0 & 94.2 & 83.4 & 89.0 & 66.8 & 75.0 & 83.6 & 90.0 & 82.2 & 87.0 & 83.3 & 89.0 \\
        \bottomrule
    \end{tabular}}
\end{table*}

We investigate whether geometry-inspired transition credit produces reliable direct reasoning. The main text tests answer quality, preservation of correct states, deployment cost, component contributions, and training dynamics; benchmark-level held-out evaluations and supplemental transfer tests appear in \Cref{app:energy_complexity}.

\subsection{Protocol}

\paragraph{Models.}

We evaluate LSPO across the Qwen3.5 family \citep{qwen3.5} to assess scaling dynamics. Qwen3.5-27B is the primary model for component and mechanism analysis; Qwen3.5-35B-A3B supplies transfer evidence for the mixture-of-experts setting alongside its architectural ablation. We also include reference metrics from gpt-oss-120b \citep{openai2025gptoss} and Claude Sonnet 4.5 \citep{anthropic2025claudesonnet45}.

\paragraph{Benchmarks.}

The evaluation spans six suites covering MATH-500 \citep{hendrycks2021measuring}, AIME25 \citep{AIME25}, GPQA \citep{rein2024gpqa}, LiveCodeBench \citep{jain2024livecodebench}, ZebraLogic \citep{lin2025zebralogic}, and MMLU-Pro \citep{wang2024mmlu}. Code tasks measure execution validity through pass@$k$ evaluation \citep{chen2021evaluating}. We additionally incorporate GSM8K \citep{cobbe2021training}, ARC-Challenge \citep{clark2018think}, BBH \citep{suzgun2023challenging}, HumanEval \citep{chen2021evaluating}, and MBPP \citep{austin2021program} to assess transfer to task families outside the training distribution.

\paragraph{Baselines.}

We compare LSPO against several baselines, including the preference optimization methods DPO \citep{rafailov2023direct} and Step-DPO \citep{lai2024step}, the RL approaches GRPO \citep{guo2025deepseek} and GDPO \citep{panaganti2026group}, and SCoRe \citep{kumar2024training}, a self-correction method that learns directly from revision traces generated by the model. Matched training alternatives for candidate selection, answer internalization, coordinate controls, and test-time reflection complement these primary baselines.

\paragraph{Implementation Details.}

All primary comparisons use fixed configurations across random seeds 42, 1234, and 3407. The primary runs use short training sequences with $T_{\max} = 6$, top-$p$ 0.95, an 8-dimensional auxiliary lift $d_z = 8$, and an internalization weight of 0.3. \Cref{app:ablation_stability} details optimizers, sampling parameters, sweep ranges, matched training alternatives, and extended analyses. The deployment comparison reports direct pass@1 generation for LSPO alongside an iterative reflection baseline evaluated with two inference passes under identical seeds and settings. Training targets use verifiers and search; the deployment comparison samples from the trained conditional model.

\subsection{Main Results}

\paragraph{Overall Performance.}

\begin{figure*}[t]
    \centering
    \includegraphics[width=.98\textwidth]{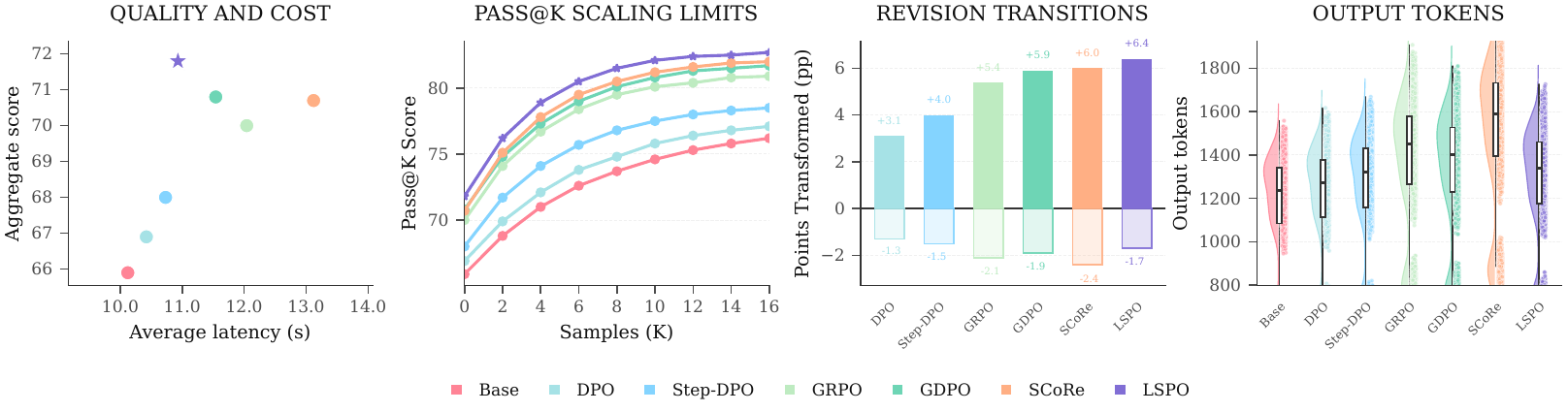}
    \caption{Quality, sampling, revision, and token cost on Qwen3.5-27B. LSPO reaches 71.8\% accuracy at 10.9\,s latency in a single pass; test-time reflection adds 0.1 points but costs 4.3\,s more and 440 additional tokens. LSPO remains above the pass@$k$ curve, most often converts incorrect answers to correct ones, and avoids the long latency tail of repeated refinement.}
    \label{fig:27b_overview}
\end{figure*}

\Cref{tab:main_results_all_models} reports direct pass@1 accuracy across six benchmark suites. LSPO achieves the highest aggregate accuracy among direct-generation baselines on each Qwen3.5 backbone, with the largest margins on structured reasoning tasks, including MATH-500, AIME25, ZebraLogic, and LiveCodeBench, where local intermediate state transitions crucially govern final outcomes. On the knowledge-oriented GPQA and MMLU-Pro benchmarks, LSPO improves over the base model while maintaining a compact direct-generation profile. The pattern holds across all three backbones in \Cref{app:full_benchmarks}, with the largest improvements at 9B and 27B where exploration has more headroom to reshape generation.

\paragraph{Deployment Efficiency.}

\Cref{fig:27b_overview} shows the LSPO policy reaching 71.8\% aggregate accuracy with 10.93\,s average latency and 1265 output tokens. It remains above the pass@$k$ curve across sampled budgets and executes the highest volume of constructive edits. Test-time reflection reaches 71.9\% with 15.18\,s and 1705 tokens across two inference passes. LSPO thus achieves comparable observed accuracy in a single pass with 4.25\,s lower latency and 440 fewer tokens per answer, serving each answer directly through the compiled generator.

\subsection{Mechanism Analyses}

\paragraph{Preserving Improvements.}

\begin{wraptable}{r}{0.39\textwidth}
    \centering
    \captionsetup{font=small}
    \vspace{-1\baselineskip}
    \caption{Preservation of improvements on Qwen3.5-27B. LSPO yields the highest net gain and lowest energy across all methods.}
    \label{tab:internalization_27b}
    \scriptsize
    \begin{tabular}{lccc}
        \toprule
        Method & Net gain & Direct $E$ & Terminal $E$ \\
        \midrule
        Base & 0.0 & 0.468 & N/A \\
        DPO & 1.8 & 0.450 & N/A \\
        Step-DPO & 2.5 & 0.438 & N/A \\
        GRPO & 3.3 & 0.420 & 0.203 \\
        GDPO & 4.0 & 0.411 & 0.201 \\
        SCoRe & 3.6 & 0.411 & 0.197 \\
        \midrule
        LSPO & 4.7 & 0.398 & 0.195 \\
        \bottomrule
    \end{tabular}
    \vspace{-2\baselineskip}
\end{wraptable}

\Cref{tab:internalization_27b} measures whether policies retain improvements found along training paths while protecting correct states. Over the same paired-answer denominator, LSPO repairs 6.4\% and damages 1.7\% of answers, yielding a 4.7-point net gain. It records the lowest direct energy of $0.398$ and terminal energy of $0.195$ among methods with an energy head; the terminal value bounds exploration capacity during training, while the direct value measures whether improvements are internalized into the policy. The span audit preserves 95.8\% of initially valid spans and yields a 16.0-point net validity gain as detailed in \Cref{app:internalization}.

\paragraph{Sources of Improvement.}

\begin{wraptable}{r}{0.39\textwidth}
    \centering
    \captionsetup{font=small}
    \vspace{-1\baselineskip}
    \caption{Training alternatives and deployment cost. Test-time reflection matches LSPO accuracy at higher deployment cost.}
    \label{tab:main_attribution}
    \scriptsize
    \setlength{\tabcolsep}{2.5pt}
    \begin{tabular}{lccc}
        \toprule
        Method & Agg. $_{\pm \text{SD}}$ & Latency (s) & Tokens \\
        \midrule
        Select then SFT & 70.6 $_{\pm 0.17}$ & 10.58 & 1218 \\
        SI only & 70.8 $_{\pm 0.16}$ & 10.74 & 1246 \\
        GRPO & 70.0 $_{\pm 0.18}$ & 12.04 & 1370 \\
        SCoRe & 70.7 $_{\pm 0.18}$ & 13.12 & 1501 \\
        Reflection & 71.9 $_{\pm 0.19}$ & 15.18 & 1705 \\
        \midrule
        LSPO & 71.8 $_{\pm 0.15}$ & 10.93 & 1265 \\
        \bottomrule
    \end{tabular}
    \vspace{-2\baselineskip}
\end{wraptable}

\Cref{tab:main_attribution} separates gains from target selection and internalization from those of the full LSPO objective, then compares the resulting policy with test-time reflection at deployment. LSPO exceeds Select then SFT and selected internalization, listed as SI only in the table, by 1.2 and 1.0 aggregate points, respectively. This reflection baseline reaches comparable accuracy at 71.9\% while incurring substantially higher deployment overhead, requiring two inference passes that add 4.25\,s latency and 440 tokens per query.

\paragraph{Generalization Across Task Families.}

Direct generation and frozen energy ordering are evaluated on five benchmark suites held out from policy training, energy fitting, and endpoint selection in \Cref{app:energy_complexity}. LSPO improves over Base on every held-out benchmark and reaches an average of 74.9, above SCoRe at 74.3. The frozen energy model also preserves local candidate ordering on this benchmark-level holdout, showing that the learned state signal remains useful on task families excluded from fitting and endpoint selection.

\paragraph{Training Dynamics.}

\begin{figure}[t]
    \centering
    \includegraphics[width=.98\linewidth]{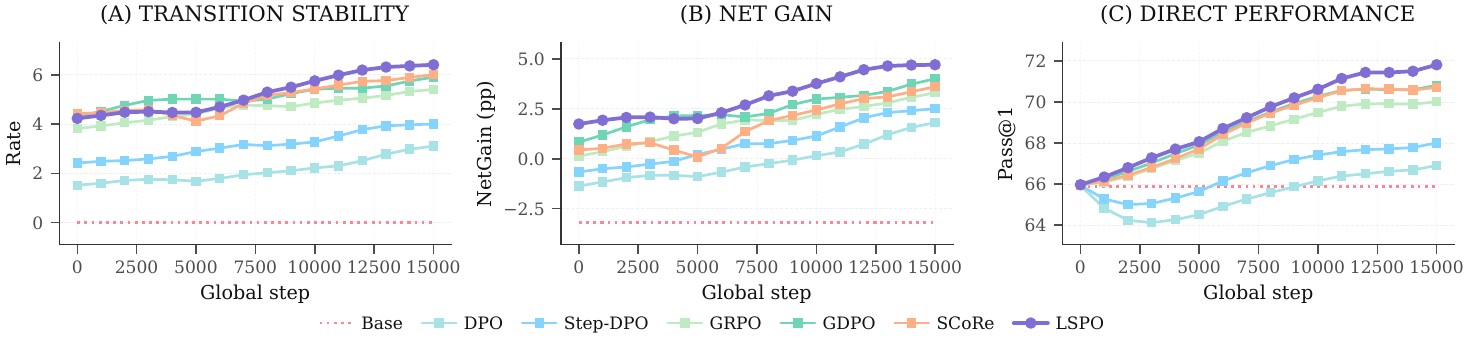}
    \caption{Training dynamics across transition stability, net gain, and direct accuracy. Transition stability and net gain plateau before direct accuracy, consistent with the geometric landscape forming before the generator absorbs it.}
    \label{fig:training_dynamics}
\end{figure}

\Cref{fig:training_dynamics} tracks the learning phases. Transition stability and net gain rise rapidly in the early steps, marking the formation of a geometric landscape before the accuracy curve plateaus. LSPO first stabilizes constructive local moves and then converts this geometry into deployed accuracy. In contrast, SCoRe trains on unconstrained self-revision traces without step penalties, which leads to aggressive rewriting of correct spans and higher C$\to$W degradation, regressing initially correct answers into erroneous states. LSPO avoids this failure mode by bounding the revision scope throughout training.

\paragraph{Component Ablations.}

\begin{wraptable}{r}{0.59\textwidth}
    \centering
    \captionsetup{font=small}
    \vspace{-1\baselineskip}
    \caption{Component ablations on Qwen3.5-27B. The full LSPO stack leads in aggregate score while controlling latency, output length, and training cost.}
    \label{tab:ablation_27b}
    \footnotesize
    \resizebox{\linewidth}{!}{%
    \begin{tabular}{lccccc}
        \toprule
        Variant & Agg. & Net gain & Latency (s) & Tokens & GPU hrs \\
        \midrule
        LSPO full & 71.8 & 4.7 & 10.93 & 1265 & 2983 \\
        - w/o lifting & 70.9 & 3.2 & 11.70 & 1316 & 2804 \\
        - w/o progress reward & 70.0 & 1.6 & 12.35 & 1366 & 2715 \\
        - w/o internalization & 70.3 & 2.4 & 12.02 & 1341 & 2566 \\
        \midrule
        GRPO & 70.0 & 3.3 & 12.04 & 1370 & 2386 \\
        DPO & 66.9 & 1.8 & 10.42 & 1196 & 1218 \\
        \bottomrule
    \end{tabular}%
    }
    \vspace{-1\baselineskip}
\end{wraptable}

\Cref{tab:ablation_27b} isolates each core mechanism. Removing the lifted coordinate reduces aggregate quality and net transition gain, while dropping the progress reward lowers net gain and inflates output verbosity. Removing answer internalization degrades retention and lowers aggregate accuracy. Parameter-matched coordinate controls retain the same training budget yet remain 0.9--1.5 aggregate points below full LSPO, supporting a role for learned refinement context beyond parameter count alone. The full LSPO stack requires additional training compute, while its structural constraints reduce deployment latency and token count relative to unshaped RL baselines. \Cref{app:ablation_stability} provides the coordinate controls, matched training alternatives, and the full MoE ablation.

\paragraph{Policy Energy and Task-Level Gains.}

\begin{figure}[t]
    \centering
    \begin{minipage}[t]{0.57\linewidth}
        \centering
        \includegraphics[width=\linewidth]{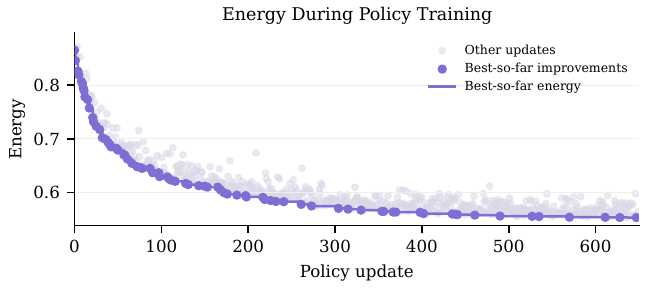}
    \end{minipage}\hspace{0.008\linewidth}
    \begin{minipage}[t]{0.41\linewidth}
        \centering
        \includegraphics[width=\linewidth,trim={2pt 4pt 1pt 0pt},clip]{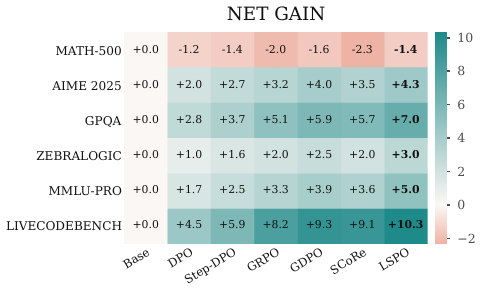}
    \end{minipage}
\caption{Policy-training energy and task-level net gains. The left panel shows recorded energy across policy updates, where the curve tracks the cumulative minimum and points mark new minima. The right panel independently reports net correction gains across benchmarks.}
    \label{fig:lspo_geometric_descent}
\end{figure}

\Cref{fig:lspo_geometric_descent} tracks recorded energy across policy updates alongside task-level correction outcomes. Energy decreases rapidly during early training and then approaches a plateau, with fluctuations between updates. The cumulative minimum curve records the lowest energy observed across updates, with highlighted points marking new minima. The right panel reports benchmark-level net correction gains independently, with LSPO showing negative net gain on MATH-500 and positive net gains on the other five benchmarks.

\section{Discussion}
\label{sec:discussion}

\paragraph{Scope and Boundaries of Local Correction.}

The transition audits show positive net correction gains on five of six benchmarks, with MATH-500 marking an operational boundary for test-time revision. In tightly coupled algebraic derivations, modifying an isolated step without revising preceding assertions leaves upstream inconsistencies that cause test-time perturbations to degrade more answers than they repair across all methods. As illustrated in \Cref{app:qualitative} on AIME25, resolution requires rebuilding from an alternative decomposition beyond finite-radius edits. In this regime, LSPO internalizes and preserves valid proof structures; training on low-energy trajectories commits coherent derivations to the generator, allowing direct accuracy in \Cref{tab:main_results_all_models} to advance even where test-time trajectory gain does not. Removing internalization costs 1.5 aggregate points and degrades correct-span retention, confirming that the generator drives this gain.

Prior analyses of iterative self-correction \citep{madaan2023self, shinn2023reflexion, huang2023large, ma2025s2r} describe the same distinction. Local correction repairs errors that represent isolated adjustments within a sound plan, including sign choices, index offsets, and lemma applications. Conversely, when the overarching plan requires fundamental restructuring, local energy gradients offer no direct descent toward the target answer. Traversing such global transitions requires either test-time search or training curricula that expose the policy to plan-level pivots.

\paragraph{Training Cost and Deployment Trade-offs.}

LSPO invests 2983 GPU-hours during training to yield per-query reductions of 4.25\,s in latency and 440 output tokens at deployment. The return on this compute investment scales with serving volume, where low-throughput regimes favor minimal training overhead, while high-throughput serving amortizes it through cumulative per-query latency and token savings.

This compute allocation also differs qualitatively from test-time scaling. Reflection-based methods \citep{snell2024scaling, zhang2025survey, sun2026reinforcement} let a single checkpoint serve both simple and complex queries by dynamically varying inference compute. In contrast, LSPO compiles deliberation directly into network parameters, producing a faster single-pass deployment where each query incurs identical inference cost regardless of difficulty. Hybrid designs that use LSPO as a fast initial pass and fall back to explicit search on high-energy outputs could combine the advantages of both regimes.

\paragraph{Representational Role of the Lifted Coordinate.}

Frozen probing evaluations demonstrate that appending the learned coordinate $z$ to textual representations improves action prediction and energy ranking on held-out transitions in \Cref{tab:lifted_coordinate_probe}, whereas prompt-shuffled controls degrade performance, confirming that the coordinate carries prompt-specific information inaccessible to a text predictor. Architecturally, the coordinate can encode compressed revision histories reflecting edit count and convergence, or establish distinct neighborhoods through Switch to identify candidate answer basins, both of which match empirical observations.

The learned coordinate guides local trajectory steps within candidate vicinities. As shown by the selection-budget analysis in \Cref{app:energy_complexity}, energy-based ranking discriminates reliably across moderate candidate neighborhoods, whereas unconstrained candidate expansion encounters non-convex surface irregularities. The lifted coordinate thereby provides directionality along local improvement paths, guiding policy updates without assuming global convexity.

\section{Related Work}
\label{sec:related}


\paragraph{Final Answers, Steps, and Trajectories.}

Outcome supervision and preference learning supply signals over final answers through RLHF \citep{christiano2017deep, ziegler2019fine, stiennon2020learning}, constitutional AI \citep{bai2022constitutional}, and direct optimization objectives \citep{rafailov2023direct}. ORPO \citep{hong2024orpo}, SimPO \citep{meng2024simpo}, and rewards generated by the model \citep{yuan2024self} refine this interface for ranking answers \citep{liu2025survey}. Process supervision moves the learning signal into intermediate reasoning steps \citep{cobbe2021training, uesato2022solving, lightman2023let}, with Step-DPO as a prominent example \citep{lai2024step}. Unlike absolute step scores suited to reranking individual states, LSPO credits relative transitions net of edit and step costs within a prompt's candidate pool, steering policy movement rather than global state ranking. GRPO \citep{guo2025deepseek} and GDPO \citep{panaganti2026group} optimize returns over sampled trajectories.

\paragraph{Learning from Refinement.}

Refinement during inference enables models to critique, revise, or retry initial answers \citep{madaan2023self, shinn2023reflexion, huang2023large, sun2026reinforcement}. Recent training variants convert these revisions into supervision through rewards generated by the model, revision analysis, and refinement pipelines for code \citep{liu2024large, ma2025s2r, bensal2025reflect, xiong2025self, jiang2026reflexicoder}. SCoRe is the closest baseline, learning from refinement traces generated by the model \citep{kumar2024training}.

\paragraph{Representing Intermediate States.}

State representation determines which movements can be compared and optimized. Plain text exposes the current answer but confines candidate alternatives to a brittle discrete space. Calibration and semantic-uncertainty studies show that confidence and correctness signals reside on separable manifolds in the hidden state \citep{guo2017calibration, kuhn2023semantic, farquhar2024detecting}; semantic probes and over-confidence analyses further confirm that correctness structure is geometrically accessible \citep{kossen2024semantic, morosini2026too}. Complementary geometric methods treat latent representations as substrates for reasoning and model connectivity \citep{xie2025mhc, zhuang2026geometric}. LSPO operationalizes this view by lifting surface answers into an augmented state space, ordering candidate transitions by energy, and transferring the selected lower-energy answer into direct generation.

\section{Limitations and Future Directions}
\label{sec:limitations}

The framework is designed for verifiable reasoning tasks where programmatic checkers or scalar outcomes ground the energy landscape. Policy improvement relies on alignment between the learned energy $\tilde E(s;q)$ and true answer validity. While intermediate fluctuations cancel telescopically along trajectories, terminal discrimination remains bounded by the fidelity of the fitted scoring head. Operating over visited neighborhoods, transition credit addresses local repairs within an established derivation, while problems requiring global proof reformulations from scratch remain outside this local search scope. Trajectory exploration concentrates compute during training, making the framework most advantageous for high-throughput serving environments where amortized efficiency provides operational gains.

Our analysis evaluates scaling across the Qwen3.5 family \citep{qwen3.5}, the 35B-A3B mixture-of-experts backbone, and cross-family transfer on Gemma-4-E4B-it \citep{google2026gemma4e4bit}. In this work, the auxiliary coordinate models textual deliberation; incorporating tool executions, evidence graphs, or multi-turn interaction traces would define the corresponding state representation for structured environments. Preliminary evaluations in \Cref{app:energy_complexity} show positive policy transfer on open-form writing and roleplay benchmarks, with geometric ordering aligning best with instruction fit, coherence, and character consistency. More comprehensive open-ended creative tasks remain to be evaluated under distinct criteria. Further advances will require latent state spaces that score and retain structured movements across these varied task formats.

\section{Conclusion}
LSPO establishes that direct policy execution can attain the reliability of iterative refinement. It matches the accuracy of test-time reflection in a single pass, and transition diagnostics confirm that the policy repairs incorrect states while preserving correct ones. The learned stopping behavior mitigates redundant revisions, and the policy transfers across model scales, to held-out benchmarks, and to mixture-of-experts architectures. Attributing credit to geometric transitions and internalizing selected endpoints thus relocates deliberation from inference into the deployed policy.

\begin{ack}
We thank the anonymous reviewers for their constructive feedback. This work was supported by the School of Information Science and Technology at Hangzhou Normal University. The authors declare no competing interests.
\end{ack}


\small
\bibliography{refs}
\bibliographystyle{plainnat}

\clearpage
\appendix
\normalsize
\section*{Appendices}

The appendix follows the evidential order of the main text. It first fixes the formal objects used by LSPO, then tests scale and transfer, isolates training mechanisms, audits internalization and energy ordering, and closes with cases that mark the method boundary.
\begin{itemize}
  \item \Cref{app:theory_roadmap} derives the path form with KL regularization and the information criterion for lifted states.
  \item \Cref{app:full_benchmarks} reports scale and transfer across Qwen sizes, benchmark families, and a distinct Gemma backbone.
  \item \Cref{app:ablation_stability} connects the update contract to component ablations, sensitivity analyses, and the fixed training configuration.
  \item \Cref{app:internalization} audits alignment from training to direct generation, stopping behavior, and the boundary with test-time reflection.
  \item \Cref{app:energy_complexity} quantifies deployment cost, energy ordering on benchmark suites held out from training, and compute accounting.
  \item \Cref{app:qualitative} maps the mechanism onto representative repairs and a global search failure case.
\end{itemize}

\small
\counterwithin{figure}{section}
\counterwithin{table}{section}
\counterwithin{equation}{section}

\section{Formal Properties of the Path Objective and State Lifting}
\label{app:theory_roadmap}

This appendix studies an unrestricted path-distribution relaxation and an information criterion for lifting. Their conclusions concern the stated distributions and assumptions.

\begin{proposition}[Gibbs form of the path optimizer constrained by KL divergence]
\label{prop:gibbs_path_distribution}
Fix a prompt $q$, let $P_{\mathrm{ref}}(\cdot \mid q)$ be a reference distribution over finite correction trajectories, and define the path score
\begin{equation}
    F_q(\tau) = \sum_{t=0}^{T(\tau)-1} r_t.
\end{equation}
Consider the variational problem
\begin{equation}
    \sup_{P \ll P_{\mathrm{ref}}(\cdot \mid q)}
    \left\{
        \mathbb{E}_{\tau \sim P}[F_q(\tau)]
        -
        \beta D_{\mathrm{KL}}\!\big(P \,\|\, P_{\mathrm{ref}}(\cdot \mid q)\big)
    \right\}.
\end{equation}
For $\beta>0$ and $0<Z_q<\infty$, its unique maximizer is the exponentially tilted path distribution
\begin{equation}
    \frac{dP_q^\star}{dP_{\mathrm{ref}}}(\tau)
    =
    \frac{\exp(F_q(\tau)/\beta)}{Z_q},
    \qquad
    Z_q
    =
    \mathbb{E}_{\tau \sim P_{\mathrm{ref}}(\cdot \mid q)}
    \left[\exp(F_q(\tau)/\beta)\right],
    \label{eq:tilted_path}
\end{equation}
and the optimal objective value equals $\beta \log Z_q$.
\end{proposition}

\paragraph{Variational argument.}
Let $P_q^\star$ be the tilted path distribution defined in \cref{eq:tilted_path}. For any admissible $P$, we have
\begin{equation}
    \frac{dP_q^\star}{dP_{\mathrm{ref}}}(\tau)
    =
    \frac{\exp(F_q(\tau)/\beta)}{Z_q}.
\end{equation}
This gives
\begin{equation}
    \log \frac{dP_q^\star}{dP_{\mathrm{ref}}}(\tau)
    =
    \frac{F_q(\tau)}{\beta} - \log Z_q.
\end{equation}
Using
\begin{equation}
    D_{\mathrm{KL}}(P \,\|\, P_q^\star)
    =
    \mathbb{E}_{\tau \sim P}
    \left[
        \log \frac{dP}{dP_q^\star}(\tau)
    \right]
    =
    \mathbb{E}_{\tau \sim P}
    \left[
        \log \frac{dP}{dP_{\mathrm{ref}}}(\tau)
        - \frac{F_q(\tau)}{\beta}
        + \log Z_q
    \right],
\end{equation}
we rearrange to obtain
\begin{equation}
    \mathbb{E}_{\tau \sim P}[F_q(\tau)]
    -
    \beta D_{\mathrm{KL}}\!\big(P \,\|\, P_{\mathrm{ref}}(\cdot \mid q)\big)
    =
    \beta \log Z_q
    -
    \beta D_{\mathrm{KL}}(P \,\|\, P_q^\star).
\end{equation}
The right side is maximized uniquely at $P=P_q^\star$, where the KL term vanishes.

\paragraph{Path Variational Form.}
\Cref{prop:gibbs_path_distribution} casts the path problem with a KL constraint as a variational optimization whose solution tilts the reference distribution toward paths with larger cumulative progress. Together with \cref{prop:reward_decomposition}, the result places lower final energy, edit cost, and step cost within a single path preference.

\paragraph{Relation to the Implemented Regularizer.}
For a fixed prompt, a policy-induced path distribution factors as
\begin{equation}
P_\theta(\tau\mid q)=\rho_\theta(s_0\mid q)
\prod_t\pi_\theta(a_t\mid s_t,q)K_\theta(s_{t+1}\mid s_t,a_t,q),
\end{equation}
including the Stop decision and absorbing termination. Under absolute continuity, the chain rule gives
\begin{equation}
\begin{aligned}
D_{\mathrm{KL}}(P_\theta\|P_{\mathrm{ref}})
={}&D_{\mathrm{KL}}(\rho_\theta\|\rho_{\mathrm{ref}})\\
&+\mathbb E_{P_\theta}\sum_t D_{\mathrm{KL}}(\pi_\theta(\cdot\mid s_t,q)\|\pi_{\mathrm{ref}}(\cdot\mid s_t,q))\\
&+\mathbb E_{P_\theta}\sum_t D_{\mathrm{KL}}(K_\theta(\cdot\mid s_t,a_t,q)\|K_{\mathrm{ref}}(\cdot\mid s_t,a_t,q)).
\end{aligned}
\end{equation}
Equality with the action-KL term requires shared initial-state distributions and conditional transition kernels. Within update $k$, LSPO fixes $\rho_{\bar\theta_k}$ and $K_{\bar\theta_k}$ for both the controller and the reference, whereby the first and third terms above cancel and the path KL reduces to the visited-state action KL. Controller RL therefore targets the fixed-mechanism controller objective in \cref{eq:rl_objective} rather than the unrestricted path objective; the snapshot is refreshed at the next update. The Gibbs result concerns the unrestricted path-distribution relaxation, whose optimum further requires realizability under the specified dynamics.

\subsection{Information in the Lifted State}
\label{sec:appendix_lifting_information}

The lifted coordinate $z_t$ augments the discrete textual state. Whether lifting adds value depends on the information it carries about constructive moves beyond what the surface text and prompt alone provide.

\begin{proposition}[Lifted state reduces optimal move prediction loss by the conditional mutual information]
\label{prop:lifting_cmi}
Let $U_t$ denote any target associated with state $s_t$, for example an indicator that a candidate local move decreases energy without harming correctness. Define the Bayes optimal predictors
\begin{equation}
    p^\star_{\mathrm{surf}}(u \mid x, q)
    =
    \mathbb{P}(U_t=u \mid X_t=x, Q=q)
\end{equation}
and
\begin{equation}
    p^\star_{\mathrm{lift}}(u \mid x, z, q)
    =
    \mathbb{P}(U_t=u \mid X_t=x, Z_t=z, Q=q).
\end{equation}
The achievable reduction in log loss from using the lifted state instead of the surface state alone is
\begin{equation}
    \mathbb{E}\!\left[-\log p^\star_{\mathrm{surf}}(U_t \mid X_t, Q)\right]
    -
    \mathbb{E}\!\left[-\log p^\star_{\mathrm{lift}}(U_t \mid X_t, Z_t, Q)\right]
    =
    I(U_t; Z_t \mid X_t, Q).
\end{equation}
\end{proposition}

\paragraph{Information identity.}
By the definition of Bayes optimal log loss,
\begin{equation}
    \mathbb{E}\!\left[-\log p^\star_{\mathrm{surf}}(U_t \mid X_t, Q)\right]
    =
    H(U_t \mid X_t, Q)
\end{equation}
and
\begin{equation}
    \mathbb{E}\!\left[-\log p^\star_{\mathrm{lift}}(U_t \mid X_t, Z_t, Q)\right]
    =
    H(U_t \mid X_t, Z_t, Q).
\end{equation}
Subtracting yields
\begin{equation}
    H(U_t \mid X_t, Q) - H(U_t \mid X_t, Z_t, Q)
    =
    I(U_t; Z_t \mid X_t, Q).
\end{equation}
This is the conditional mutual information identity.

\paragraph{Lift Criterion.}
For a fixed joint distribution of $(Q,X_t,Z_t,U_t)$ with finite conditional entropies, the identity yields an information criterion. The lifted coordinate $z_t$ improves prediction exactly when it carries information about constructive moves absent from the surface text and input. Positive conditional mutual information equals the achievable reduction in optimal log loss.

For a deterministic reprojection $Z_t=g(Q,X_t)$ at fixed parameters, the conditional mutual information is zero. Recursive coordinates can retain history since their most recent reset; positive conditional information depends on the resulting state distribution and target $U_t$. The finite-capacity probes quantify accessible predictive information, and their NLL differences give empirical prediction gains.

\section{Full Scaling Profiles and Cross-Architecture Transfer}
\label{app:full_benchmarks}

This appendix extends the 27B analysis across benchmarks, model scales, and pass@$k$ budgets. Following established protocols for reasoning and executable code evaluation \citep{wang2022self, chen2021evaluating}, it reports performance for each benchmark across all three Qwen backbones.

\begin{figure*}[t]
    \centering
    \includegraphics[width=.94\textwidth]{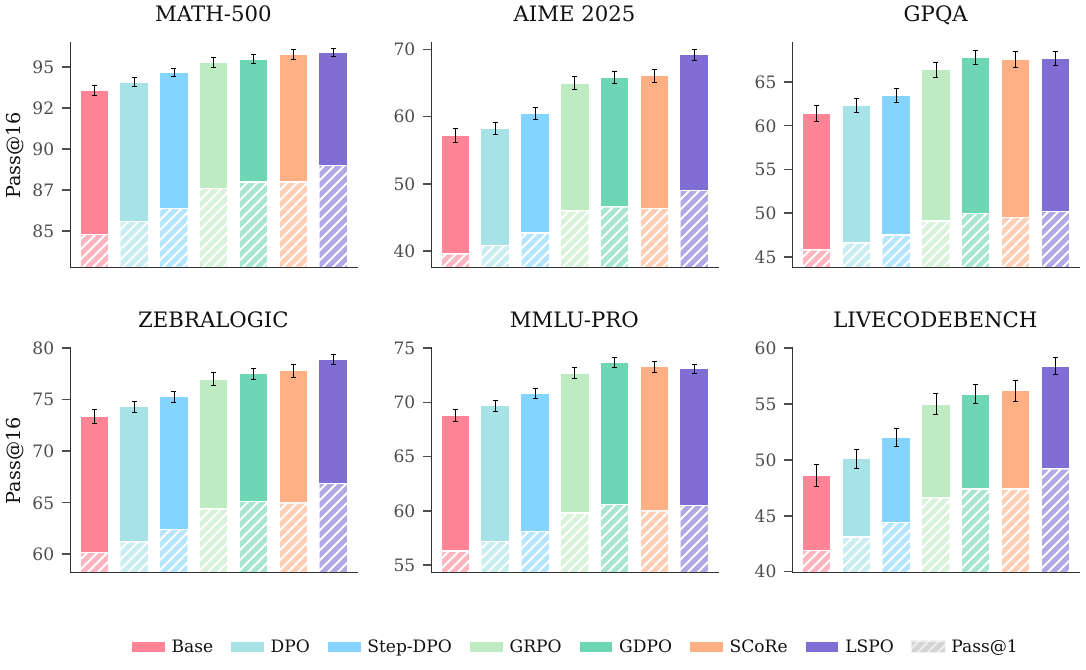}
    \caption{Benchmark profile for Qwen3.5-9B. LSPO improves direct accuracy while preserving substantial sampled pass@16 headroom in the smaller dense backbone.}
    \label{fig:appendix_pass16_9b}
\end{figure*}

\begin{figure*}[t]
    \centering
    \includegraphics[width=.94\textwidth]{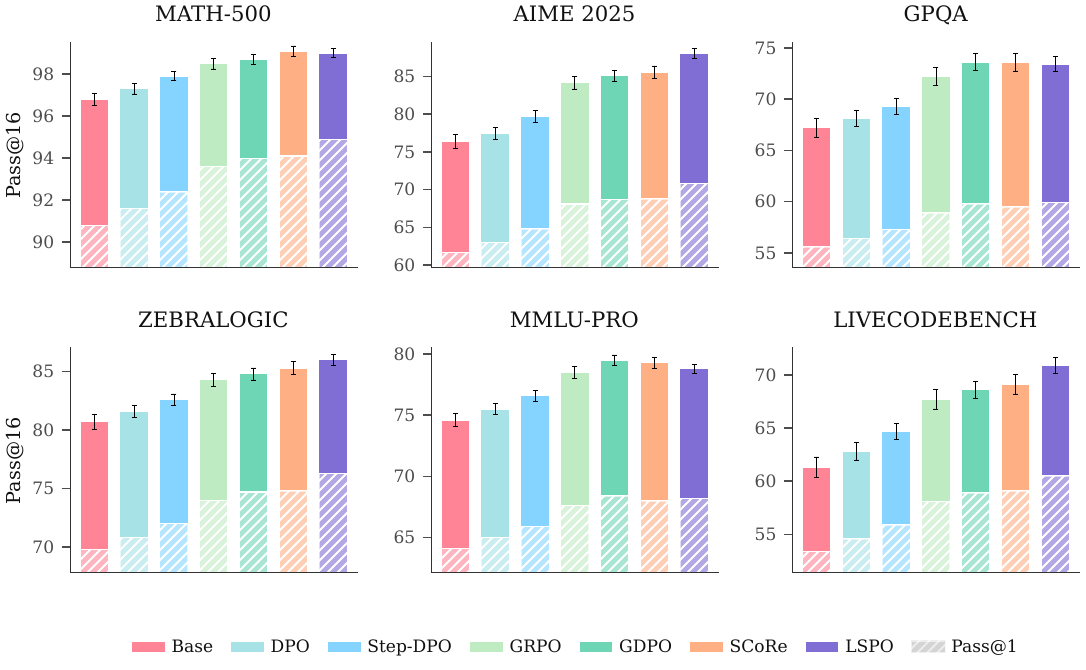}
    \caption{Benchmark profile for Qwen3.5-27B. LSPO achieves consistent direct gains while preserving the sampled pass@16 envelope.}
    \label{fig:appendix_pass16_27b}
\end{figure*}

\begin{figure*}[t]
    \centering
    \includegraphics[width=.94\textwidth]{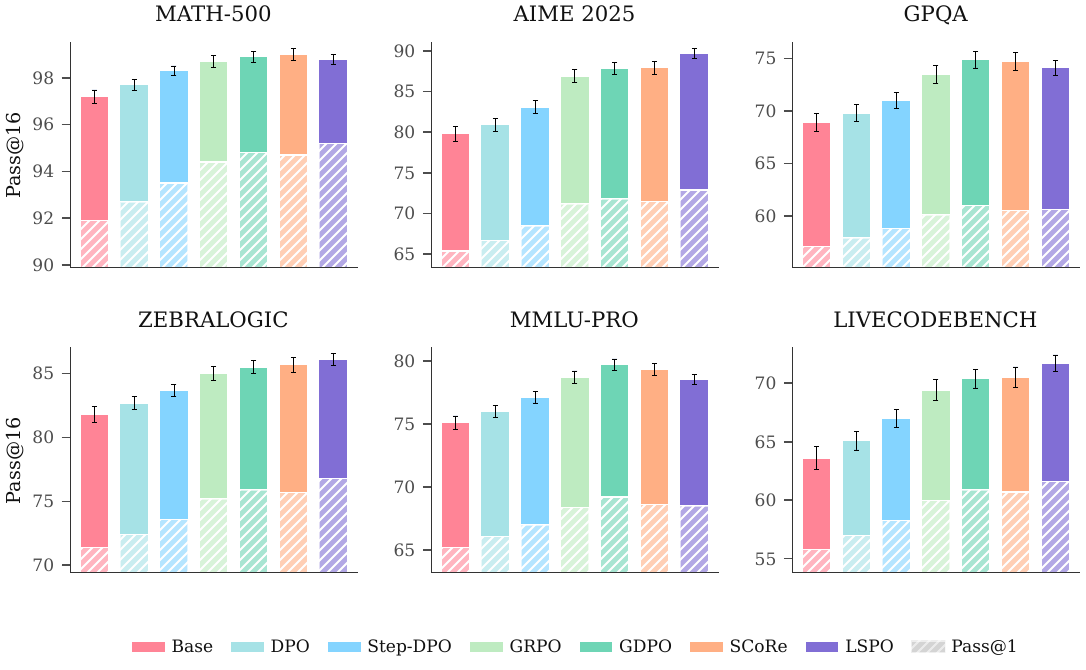}
    \caption{Benchmark profile for Qwen3.5-35B-A3B. LSPO remains the strongest direct policy as available performance headroom narrows in the MoE backbone.}
    \label{fig:appendix_pass16_35b}
\end{figure*}

Across scales, LSPO attains the highest direct accuracy without collapsing the pass@16 envelope, while expanding sampled ceilings on complex reasoning suites, as illustrated in \Cref{fig:appendix_pass16_9b,fig:appendix_pass16_27b,fig:appendix_pass16_35b}. The separation is largest on MATH-500, GPQA, and ZebraLogic, where methods based on preferences capture less of the available revision signal than on broad knowledge tasks. Smaller models retain wider gaps between pass@1 and pass@16, indicating more residual search headroom.

These distributions relate LSPO to recent reasoning systems operating at scale, such as DeepSeekMath and DeepSeek-R1 \citep{shao2024deepseekmath, guo2025deepseek}. Reasoning RL at large scale can improve quality by reshaping output distributions and spending inference compute. The pass@16 diagnostics isolate the portion of that improvement absorbed into deployed generation while sampled diversity remains available.

\subsection{Transfer Across Backbones}

\begin{table}[t]
    \centering
    \caption{Transfer across backbones to Gemma-4-E4B-it. LSPO attains the highest aggregate score on this distinct architecture while incurring only a minor latency overhead relative to direct preference baselines.}
    \label{tab:cross_family_check}
    \small
    \begin{tabular}{lccccc}
        \toprule
        Method & MATH-500 & GPQA & LCB & Aggregate & Latency (s) \\
        \midrule
        Base & 87.6 & 48.7 & 47.1 & 61.1 & 6.81 \\
        DPO & 88.3 & 49.4 & 48.0 & 61.9 & 7.04 \\
        GRPO & 89.2 & 50.1 & 49.4 & 62.9 & 7.18 \\
        \midrule
        LSPO & 90.7 & 51.3 & 52.0 & 64.7 & 7.31 \\
        \bottomrule
    \end{tabular}
\end{table}

While primary scaling experiments evaluate the Qwen3.5 family under consistent formatting, \Cref{tab:cross_family_check} examines transfer to Gemma-4-E4B-it \citep{google2026gemma4e4bit}. On this distinct backbone, LSPO attains the highest aggregate score, improving over both offline preference optimization and online reinforcement learning with only a modest latency increase. This confirms that transition credit and answer internalization transfer robustly beyond the primary Qwen family.

\subsection{Benchmark Coverage}

The benchmark suite spans complementary reasoning regimes. MATH-500 and AIME assess multi-step mathematical derivations. LiveCodeBench tests executable code synthesis under formal test specifications. ZebraLogic probes constrained deduction and local backtrack revisions. GPQA and MMLU-Pro assess multi-step scientific and domain knowledge. In addition, held-out task suites such as BBH, GSM8K, and HumanEval are evaluated in \Cref{app:energy_complexity} to audit out-of-distribution calibration.

\section{Update Mechanics, Ablations, and Sensitivity Analyses}
\label{app:ablation_stability}

This appendix links the training procedure to the mechanisms tested in the main text. The three core mechanisms of LSPO play complementary roles. Lifting defines the state in which correction moves are compared; the progress reward supplies dense local credit, aligned with reward shaping and process supervision \citep{ng1999policy, uesato2022solving, lightman2023let}; and answer internalization moves selected candidates into the deployed policy.

Baseline comparisons place LSPO mechanics against neighboring training strategies. Step-DPO \citep{lai2024step} strengthens supervision over long reasoning chains relative to standard DPO while retaining a preference objective. GDPO \citep{panaganti2026group} addresses stability in RL over sampled paths through group weighting. SCoRe \citep{kumar2024training} trains online self-correction from the model's own generation traces. LSPO combines cost-aware move shaping with answer internalization so that selected states enter generation in a single pass.

The accompanying ablations examine the contributions of state lifting and reward shaping under the reported configurations, complementing studies of shaped training signals in reinforcement learning \citep{kaufmann2024survey, lambert2025reinforcement, wu2026art}.

\subsection{Update Mechanics}

\begin{figure*}[t]
    \centering
    \includegraphics[width=\textwidth]{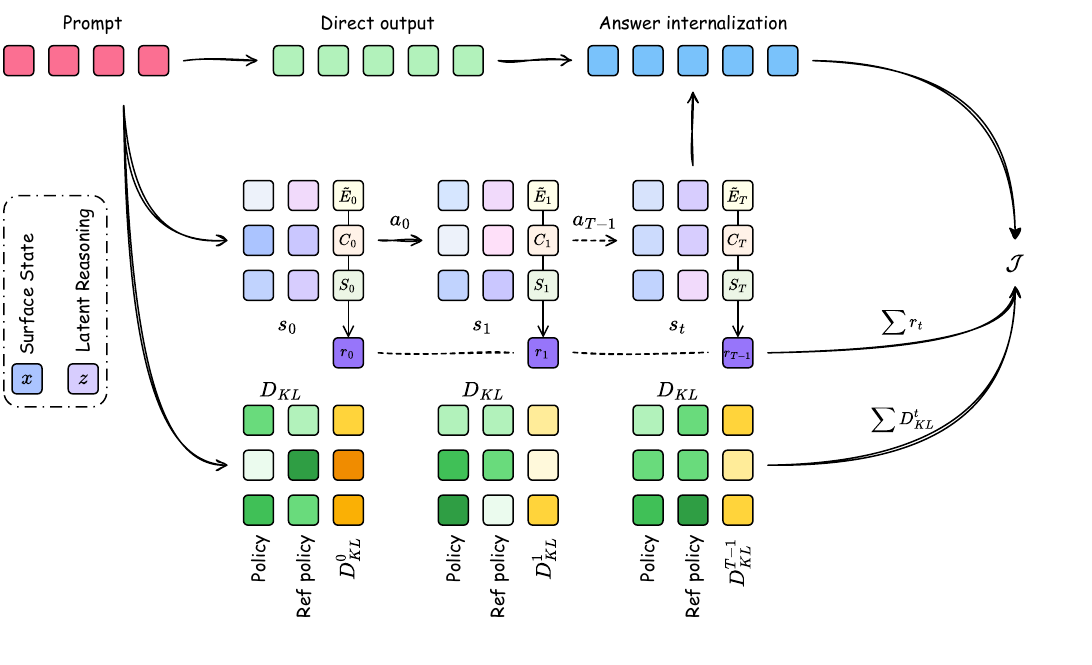}
    \caption{A single LSPO training update. Move shaping in the lifted space and answer internalization act concurrently on the selected candidate.}
    \label{fig:appendix_training_flow}
\end{figure*}

\Cref{fig:appendix_training_flow} and \Cref{alg:lspo_detailed} specify the computational graph and algorithmic procedure for a single LSPO update.

The initial generation at the top of the update graph represents the deployed answer before revision. This output initializes the lifted path and the candidate set. Each state encodes the surface text, the lifted coordinate, and the scalar energy. Each action proposes a local move and receives the reward increment from \cref{eq:progress_reward}. This decomposition combines credit assignment over paths with direct training on the selected answer. The advantage estimator targets the controller objective in \cref{eq:rl_objective}.

\subsection{Architecture and Data Transparency}

The primary Qwen3.5-27B implementation uses the architecture and public corpus mixture in \Cref{tab:architecture_specifications,tab:training_corpus}. These details fix how the lifted coordinate enters the model, how the energy head receives text and state features, and which data mixture supplies the shared training budget.

\begin{algorithm}[t]
    \caption{LSPO update contract}
    \label{alg:lspo_detailed}
    \small
    \begin{NoHyper}
    \begin{algorithmic}
        \Require minibatch $\mathcal B$; direct generator $p_\theta$; action policy $\pi_\theta$; value head $V_\omega$; reference $\pi_{\mathrm{ref}}$
        \Require frozen $(h_\xi,g_\psi,f_\eta,E_\phi)$ defining $\tilde E$; optional verifier $V$; horizon $T_{\max}$
        \Require weights $(\alpha,\beta,\lambda_{\mathrm{edit}},\lambda_{\mathrm{step}})$; optimizer $\mathrm{Opt}$
        \Ensure updated parameters $(\theta,\omega)$; deployment uses only $p_\theta(\cdot\mid q)$
        \State Fix rollout snapshot $\bar\theta_k\gets\theta$ for initial answers and text proposals
        \State Initialize path buffer $\mathcal R \gets \emptyset$ and imitation buffer $\mathcal I \gets \emptyset$
        \For{each prompt $q \in \mathcal B$}
            \State Sample direct answer $x^{\mathrm{dir}} \sim p_{\bar\theta_k}(\cdot \mid q)$
            \State Set $h_0=h_\xi(q,x^{\mathrm{dir}})$, $s_0=(x^{\mathrm{dir}},g_\psi(h_0))$, and $\mathcal C \gets \{(x^{\mathrm{dir}},s_0,\tilde E(s_0;q))\}$
            \For{$t=0,\ldots,T_{\max}-1$}
                \State Sample $a_t \sim \pi_{\bar\theta_k}(\cdot\mid s_t,q)$ from \{\texttt{revise}, \texttt{switch}, \texttt{stop}\}
                \If{$a_t=\texttt{stop}$}
                    \State Set $r_t=0$, $s_{t+1}=s_t$, and $\mathrm{done}=\mathrm{True}$
                    \State Record $(q,s_t,a_t,s_t,0,\log\pi_{\bar\theta_k}(a_t\mid s_t,q),\log\pi_{\mathrm{ref}}(a_t\mid s_t,q),\mathrm{done})$ in $\mathcal R$
                    \State \textbf{break}
                \EndIf
                \State Execute $a_t$ using $K_{\bar\theta_k}$ to obtain $x_{t+1}$; set $h_{t+1}=h_\xi(q,x_{t+1})$
                \State Set $b_t=z_t$ for Revise, or $b_t=g_\psi(h_{t+1})$ for Switch
                \State Set $z_{t+1}=\mathrm{LN}(b_t+f_\eta(h_{t+1}-h_t,b_t,e(a_t)))$
                \State Set $s_{t+1}=(x_{t+1},z_{t+1})$ and $\mathrm{done}=(t=T_{\max}-1)$
                \State Compute $r_t=\tilde E(s_t;q)-\tilde E(s_{t+1};q)-\lambda_{\mathrm{edit}}d(x_{t+1},x_t)-\lambda_{\mathrm{step}}$
                \State Record $(q,s_t,a_t,s_{t+1},r_t,\log\pi_{\bar\theta_k}(a_t\mid s_t,q),\log\pi_{\mathrm{ref}}(a_t\mid s_t,q),\mathrm{done})$ in $\mathcal R$
                \State Append $(x_{t+1},s_{t+1},\tilde E(s_{t+1};q))$ to $\mathcal C$
            \EndFor
            \State Select $x_T^*(q)$ from $\mathcal C$ by minimum energy, using $V$ only to break training ties
            \State Append $(q,x_T^*(q))$ to $\mathcal I$
        \EndFor
        \State Estimate advantages from $\mathcal R$ and form the sampled controller loss $\mathcal L_{\mathrm{ctrl}}^{(k)}$ with action KL
        \State Form answer internalization loss $\mathcal L_{\mathrm{SI}}=-\mathbb E_{\mathcal I}[\log p_\theta(x_T^*(q)\mid q)]$
        \State Update policy/generator with $\mathcal L_{\mathrm{ctrl}}^{(k)}+\alpha\mathcal L_{\mathrm{SI}}$; fit value head $V_\omega$ to returns
    \end{algorithmic}
    \end{NoHyper}
\end{algorithm}

\Cref{alg:lspo_detailed} fixes the training contract. The model generates the deployed answer, explores lifted states during training, selects a candidate, and updates the direct generator on that answer. The energy model, verifier, candidate selection, and explicit revision actions serve as instruments during training only.

\begin{table}[t]
    \centering
    \caption{Architectural specifications for the primary Qwen3.5-27B LSPO implementation.}
    \label{tab:architecture_specifications}
    \small
    \begin{tabularx}{\linewidth}{l>{\raggedright\arraybackslash}X>{\raggedright\arraybackslash}X>{\raggedright\arraybackslash}X}
        \toprule
        Component & Architecture & Input / output & Key setting \\
        \midrule
        Lift projection $g_\psi$ & 2-layer MLP, width 1024, GELU & Hidden state $\rightarrow z \in \mathbb{R}^{8}$ & Input LayerNorm \\
        Energy head $\tilde E$ & 3-layer MLP, width 1024, SiLU & Pooled text, $z$, task embedding $\rightarrow$ scalar & Pairwise margin 0.20 \\
        Transition block & 2 residual blocks, width 1024, SiLU & Text delta, $b_t$, action embedding $\rightarrow \Delta z$ & $z_{t+1}=\mathrm{LN}(b_t+\Delta z)$ \\
        Action embedding & Lookup table, width 128 & Revise, Switch, Stop $\rightarrow$ vector & Frozen transition conditioning \\
        Value head & 2-layer MLP, width 512, SiLU & Transition state $\rightarrow$ scalar value & GAE 0.95, clip 0.2 \\
        \bottomrule
    \end{tabularx}
\end{table}

\begin{table}[t]
    \centering
    \caption{Training corpus mixture and sample counts shared by the compared methods.}
    \label{tab:training_corpus}
    \small
    \begin{NoHyper}
    \begin{tabular}{lcc}
        \toprule
        Task family & Public corpus source & Samples \\
        \midrule
        Math & open-r1/OpenR1-Math-220k\footnotemark & 38,000 \\
        Code & BAAI/TACO\footnotemark & 24,000 \\
        Science & allenai/sciq\footnotemark & 18,000 \\
        Logic & longface/logicLM\footnotemark & 12,000 \\
        \bottomrule
    \end{tabular}
    \end{NoHyper}
\end{table}
\addtocounter{footnote}{-4}
\stepcounter{footnote}\footnotetext{\url{https://huggingface.co/datasets/open-r1/OpenR1-Math-220k}}
\stepcounter{footnote}\footnotetext{\url{https://huggingface.co/datasets/BAAI/TACO}}
\stepcounter{footnote}\footnotetext{\url{https://huggingface.co/datasets/allenai/sciq}}
\stepcounter{footnote}\footnotetext{\url{https://huggingface.co/datasets/longface/logicLM}}

\paragraph{Transition Operators.}
The action set contains three operators. Revise edits the active surface answer by sampling a bounded patch over a reasoning, equation, or code span and recursively updating its coordinate. Switch replaces the active candidate with another sampled state for the same prompt, including its surface text and projected coordinate, before the transition block applies the residual update. Stop terminates expansion and leaves the current state as a candidate for selection. The edit term $d(x_{t+1},x_t)$ is computed after normalizing whitespace, mathematical tokens, and code tokens. Long answers are segmented into reasoning spans, and the distance is the normalized Levenshtein distance at token level over changed spans, capped at one for each transition. This convention charges local rewrites without making copied context dominate the penalty.

\paragraph{Lifted Coordinate.}
Let $h_t=h_\xi(q,x_t)$ be the frozen text representation and $e(a_t)$ the frozen action embedding used by the transition block. Initialization uses $z_0=g_\psi(h_0)$. For Revise, the residual update starts from $b_t=z_t$; for Switch, it starts from $b_t=g_\psi(h_{t+1})$. Both use
\begin{equation}
z_{t+1}=\mathrm{LN}\!\left(b_t+f_\eta(h_{t+1}-h_t,b_t,e(a_t))\right).
\end{equation}
Revise therefore retains refinement history through $z_t$, while Switch resets that history to the replacement candidate. Stop preserves $z_t$. The energy $\tilde E(s_t;q)=E_\phi(h_t,z_t,q)$ and action policy $\pi_\theta(\cdot\mid s_t,q)$ consume the same stored coordinate.

\paragraph{Training Stages and Gradient Scope.}
We first fit the scoring branch on sampled training states using the energy-ranking supervision described below. Before policy optimization, we freeze the text encoder $h_\xi$, projection $g_\psi$, transition block $f_\eta$ (including its conditioning embeddings and normalization), and energy head $E_\phi$. The scoring encoder is a fixed parameter copy, separate from the trainable generator. These components remain fixed throughout rollout collection, candidate selection, RL/SI updates, and ordering audits. Coordinates can evolve recursively even though the update function is frozen.

Within each update, initial answers and text proposals are sampled using the rollout snapshot and treated as fixed environment outcomes for controller RL. Shared parameters receive gradients from both controller RL and answer internalization, which maximizes the likelihood of selected answers given the original prompt. A value-regression loss trains the value head. Rewards, stored scoring features, and selected targets are detached supervision; the ranking loss is confined to scoring-branch fitting. Thus RL/SI gradients leave all scoring-branch parameters unchanged.

\paragraph{Energy Supervision.}
Energy supervision is built from comparisons among states visited for the same prompt. Checkers for exact match, executable tests, multiple choice labels, and task verifiers identify successful states. Failed answers, states rejected by verifiers, longer states with no verifier improvement, variants with noisy energy, and controls using terminal rewards only supply negatives. The scalar head is fit with a pairwise margin objective so that verified states or states with lower cost receive lower energy than alternatives from the same prompt. The head is fit on training families, then frozen for candidate selection and for the ordering audit on held-out families in \Cref{app:energy_complexity}; GSM8K, ARC-Challenge, BBH, HumanEval, and MBPP are excluded from fitting and endpoint selection in that audit.

\paragraph{Policy Update.}
Rollouts are generated by the current policy for the current minibatch, and no replay buffer carries trajectories across optimizer updates. Advantages are computed from the shaped rewards with generalized advantage estimation using discount $\gamma=1$ and $\lambda=0.95$, followed by normalization and value clipping at $0.2$. The KL term is evaluated against the fixed reference action policy $\pi_{\mathrm{ref}}$ on the same visited states. The rollout policy $\pi_{\bar\theta_k}$ supplies the sampling probabilities for the update and is distinct from this reference. Stop is recorded with zero reward and a terminal mask, giving zero value bootstrap while retaining its policy-gradient and action-KL terms. The finite rollout horizon also terminates the episode with zero bootstrap. Scoring-branch fitting precedes these policy updates.

\paragraph{Action Statistics.}
The observed action frequencies in \Cref{tab:action_statistics} show a shift from exploration toward termination over training. Revise and Switch transitions decrease energy in most executed cases, while the lower probability of reaching $T_{\max}$ later in training is consistent with the stop action terminating local basins once further movement no longer offsets its costs.

\begin{table}[t]
    \centering
    \caption{Observed action use across LSPO training phases on Qwen3.5-27B. Revise and Switch transitions are also reported by their energy decrease rates.}
    \label{tab:action_statistics}
    \small
    \begin{tabular}{lcccc}
        \toprule
        Measure / phase & Revise (\%) & Switch (\%) & Stop (\%) & Reach $T_{\max}$ (\%) \\
        \midrule
        Overall & 49.8 & 25.9 & 24.3 & 11.8 \\
        Energy decrease rate & 70.8 & 63.4 & -- & -- \\
        Early 0--30\% & 56.1 & 30.8 & 13.1 & 21.4 \\
        Late 70--100\% & 42.6 & 20.2 & 37.2 & 5.9 \\
        \bottomrule
    \end{tabular}
\end{table}

\begin{figure}[t]
    \centering
    \includegraphics[width=\linewidth]{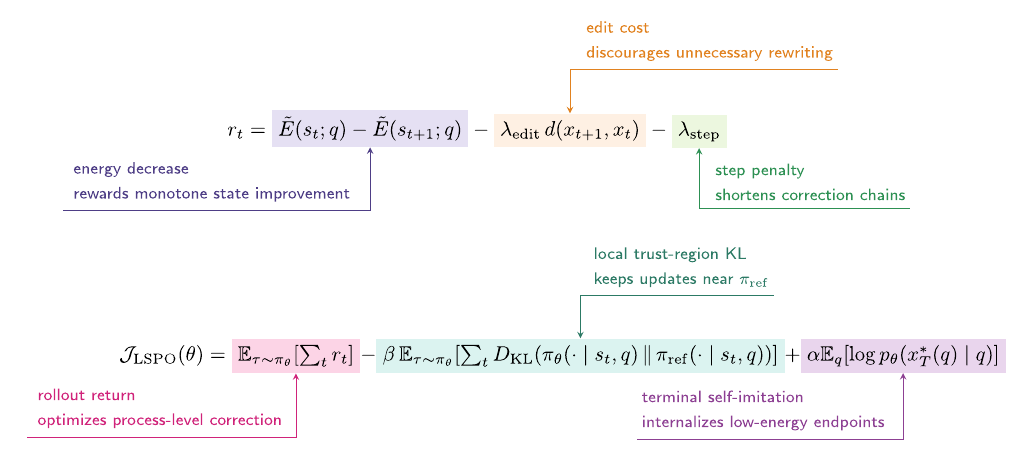}
    \caption{Annotated reward and objective for LSPO. The assignment of credit per step, the regularization of the policy, and direct training on the answer occupy distinct terms within the update.}
    \label{fig:lspo_formula_annotations}
\end{figure}

\paragraph{Objective Components.}
The flow diagram follows information propagation through the update sequence. \Cref{fig:lspo_formula_annotations} maps each symbolic block to its contribution within that update.

\subsection{Objective Context}

From a reinforcement learning perspective, LSPO addresses the balance between sparse final rewards, dense shaping signals, and variance control \citep{sutton1998reinforcement}. The progress reward gives the optimizer a local signal over lifted moves rather than raw token sequences, making constructive revisions legible while reducing the burden on credit assigned only to the final answer.

Recent reasoning RL literature reinforces the emphasis on signal structure. Reward design shapes whether policy learning remains stable and semantically aligned \citep{pan2026reward}. The LSPO ablations study the components governing state definition, local improvement criteria, and deployed execution.

Two contemporary research strands reinforce this reading. Transformed objectives improve pass@$k$ performance within frameworks similar to GRPO \citep{le2026transform}, indicating that reasoning RL is sensitive to objective design. Reward model analyses further show that downstream RL depends on accurate preferences and verifiable rationales \citep{lai2026r}. LSPO applies this lesson to individual moves, where the supervisory signal is shaped by energy descent and edit costs.

\subsection{Training Component Analysis}

\begin{figure*}[t]
    \centering
    \includegraphics[width=\textwidth]{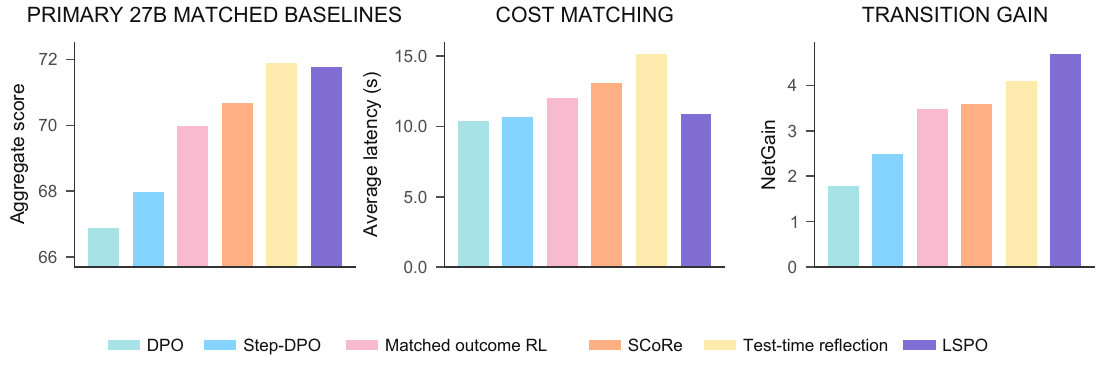}
\caption{Training alternatives on Qwen3.5-27B. LSPO achieves accuracy comparable to test-time reflection with lower deployment cost and gains from constructive state changes.}
    \label{fig:matched_27b_controls}
\end{figure*}

\begin{table}[t]
    \centering
    \caption{Matched training alternatives on Qwen3.5-27B. Evaluation examples and nominal budgets are shared; \Cref{tab:control_protocol} specifies candidate pools, reward definitions, and SI/SFT matching. SI only uses a fixed LSPO candidate pool.}
    \label{tab:matched_controls}
    \small
    \begin{tabular}{lccc}
        \toprule
        Method & Agg. $\pm$ SD & LSPO diff. (95\% CI) & Inference contract \\
        \midrule
        Select then SFT & 70.6 $\pm$ 0.17 & +1.2 [0.7, 1.6] & direct \\
        SI only & 70.8 $\pm$ 0.16 & +1.0 [0.6, 1.4] & direct \\
        Without lifting & 70.9 $\pm$ 0.18 & +0.9 [0.5, 1.3] & direct \\
        Without progress reward & 70.0 $\pm$ 0.19 & +1.8 [1.3, 2.3] & direct \\
        Return-matched terminal reward & 69.7 $\pm$ 0.20 & +2.1 [1.5, 2.6] & direct \\
        Test-time reflection & 71.9 $\pm$ 0.19 & -- & critique + revision \\
        LSPO & 71.8 $\pm$ 0.15 & -- & direct \\
        \bottomrule
    \end{tabular}
\end{table}

\begin{table}[t]
    \centering
    \caption{Extended ablations on Qwen3.5-27B. Perturbations to the lifted coordinate and corruption of energy degrade distinct metrics, separating the role of state geometry from that of local reward shaping.}
    \label{tab:appendix_extended_ablations}
    \small
    \begin{tabular}{l l c c}
        \toprule
        Family & Variant & Aggregate drop & Net gain \\
        \midrule
        Lift & $z$ dimension = 64 & 0.6 & 3.9 \\
        Lift & $z$ dimension = 2 & 0.8 & 3.7 \\
        Lift & No lift & 0.9 & 3.2 \\
        Lift & No switch & 1.4 & 3.1 \\
        Lift & Revise only & 1.7 & 2.7 \\
        Lift & Switch only & 2.0 & 2.5 \\
        Energy & No margin & 1.2 & 3.2 \\
        Energy & Noisy energy & 1.5 & 2.9 \\
        Energy & No energy stop & 1.8 & 2.4 \\
        Energy & Terminal only & 2.1 & 2.2 \\
        \bottomrule
    \end{tabular}
\end{table}

\begin{table}[t]
    \centering
    \caption{Sensitivity to energy corruption on Qwen3.5-27B. Gaussian noise weakens but preserves part of the transition gain, whereas flipped or shuffled ordering drives rank metrics toward chance.}
    \label{tab:energy_corruption_controls}
    \small
    \begin{tabular}{lccccc}
        \toprule
        Energy signal & Agg. & Drop & Net gain & AUC & Pairwise \\
        \midrule
        Clean LSPO energy & 71.8 & 0.0 & 4.7 & 0.792 & 0.734 \\
        Gaussian noise & 70.3 & 1.5 & 2.9 & 0.731 & 0.681 \\
        Label flip or shuffle & 68.6 & 3.2 & 0.7 & 0.508 & 0.506 \\
        Terminal only & 69.7 & 2.1 & 2.2 & 0.681 & 0.641 \\
        \bottomrule
    \end{tabular}
\end{table}

\begin{table}[t]
\centering
\caption{Control specifications for training alternatives. Nominal budgets specify candidate limits, update counts, and sampling settings; they do not imply identical realized token counts or GPU hours.}
\label{tab:control_protocol}
\small
\resizebox{\textwidth}{!}{%
\begin{tabular}{lcccccc}
\toprule
Variant & Candidate pool & State space & Target selector & Controller RL & Reward signal & Transfer target \\
\midrule
Full LSPO & Dynamic rollout & Lifted $(x,z)$ & Frozen $\tilde E$ & Dense policy & $\Delta \tilde E - \text{costs}$ & $\min \tilde E$ (SI) \\
Return-matched terminal reward & Dynamic rollout & Lifted $(x,z)$ & Frozen $\tilde E$ & Delayed policy & Trajectory $R(\tau)$ & $\min \tilde E$ (SI) \\
SI only on fixed pool & Fixed Full pool & Lifted $(x,z)$ & Frozen $\tilde E$ & None & None & $\min \tilde E$ (SI) \\
Select then SFT & Fixed Full pool & Surface text & Task verifier & None & None & Verifier-best (SFT) \\
Without lifting & Dynamic rollout & Surface text & Surface $E$ & Dense policy & $\Delta E - \text{costs}$ & $\min E$ (SI) \\
Without progress reward & Dynamic rollout & Lifted $(x,z)$ & Frozen $\tilde E$ & Dense policy & $-\text{costs only}$ & $\min \tilde E$ (SI) \\
\bottomrule
\end{tabular}%
}
\end{table}

\Cref{fig:matched_27b_controls} reports the primary 27B comparison. Test-time reflection reaches 71.9\% aggregate accuracy, 0.1 points above LSPO's 71.8\%, while LSPO uses 4.25\,s less latency and 440 fewer tokens. GRPO supplies a reinforcement learning comparison over final-answer rewards under the same seed set and sampling budget.

\Cref{tab:matched_controls,tab:control_protocol} compare LSPO with controlled training alternatives. SI only on a fixed LSPO candidate pool uses the candidates and selected targets produced by Full LSPO, and measures what answer internalization achieves given those teacher candidates. Select then SFT uses the same fixed pool with verifier-based selection and predetermined cost rules. Candidate construction is accounted for separately from the matched SI/SFT token exposure and updates. LSPO leads both alternatives; the paired intervals are positive. These comparisons distinguish downstream training effects conditional on the candidate pool; the teacher pool itself already reflects LSPO exploration.

\paragraph{Return-Matched Terminal Reward.}
To isolate reward timing, the terminal control retains the scoring function, edit metric, action space, target selection, termination rules, SI, and action KL of Full LSPO. For a completed trajectory with $M$ executed Revise/Switch moves and final state $s_M$, define
\begin{equation}
R(\tau;q)=\tilde E(s_0;q)-\tilde E(s_M;q)
-\lambda_{\mathrm{edit}}\sum_{t=0}^{M-1}d(x_{t+1},x_t)-\lambda_{\mathrm{step}}M.
\end{equation}
Let $L$ count all recorded decisions, including an explicit Stop when present. The terminal control pays
\begin{equation}
r_t^{\mathrm{terminal}}=
\begin{cases}
0,&t<L-1,\\
R(\tau;q),&t=L-1.
\end{cases}
\qquad
\sum_{t=0}^{L-1}r_t^{\mathrm{terminal}}
=\sum_{t=0}^{L-1}r_t^{\mathrm{dense}}=R(\tau;q).
\end{equation}
An explicit Stop receives the delayed trajectory return; at the horizon, the last executed move receives it. Immediate Stop gives $M=0$ and zero return. Full LSPO retains zero reward for Stop itself. With $\gamma=1$ and unchanged KL regularization, the two objectives assign the same return to every identical trajectory. Their sampled updates differ through reward timing, value fitting, and GAE with $\lambda=0.95$, thereby testing finite-sample optimization rather than a different ideal trajectory objective.

\Cref{tab:appendix_extended_ablations} extends the ablation analysis. Restricting the policy to paths that only revise degrades both aggregate quality and transition gain, pointing to the role of motion in the lifted state space beyond surface rewriting. \Cref{tab:energy_corruption_controls} measures sensitivity to energy corruption. Gaussian perturbations weaken but preserve part of the transition gain, while flipped or shuffled ordering drives rank metrics toward chance. The terminal reward control retains the trajectory return while delaying its feedback to termination.

\subsection{Lifted Coordinate Analysis and MoE Transfer}

The frozen probe in \Cref{tab:lifted_coordinate_probe} tests the predictive utility of $z$ for a finite-capacity predictor on held-out transitions. Text plus the learned coordinate improves action prediction, reduces NLL, and increases energy AUC; shuffling $z$ disrupts the prompt association. These probe differences measure predictive utility for the fitted probe; the information identity in \Cref{prop:lifting_cmi} concerns Bayes-optimal log loss.

\Cref{tab:coordinate_controls} evaluates whether the lifted-coordinate result is explained by parameter count or generic conditioning capacity. The controls use the same Qwen3.5-27B backbone, candidate budget, training budget, and three-seed protocol as full LSPO. A same-parameter no-lift MLP, a random coordinate, a learned nuisance coordinate, and a frozen or random projection all remain below full LSPO. Their net gains range from 2.8 to 3.0 points, compared with 4.7 for full LSPO; the paired intervals for the aggregate gaps remain positive. These results support the interpretation that the learned, prompt-conditioned coordinate carries refinement-relevant information beyond extra parameters alone.

\Cref{tab:moe_ablation} repeats the component analysis on Qwen3.5-35B-A3B. The same ordering of full LSPO over the component removals provides transfer evidence for the MoE backbone, while Qwen3.5-27B remains the setting used for the primary causal comparison.

\begin{table}[t]
    \centering
    \caption{Frozen probe for information in the lifted coordinate on transitions held out from fitting. Adding the learned coordinate improves action prediction and energy ordering; shuffled coordinates remove the association.}
    \label{tab:lifted_coordinate_probe}
    \small
    \begin{tabular}{lccc}
        \toprule
        Input condition & Action acc. $\pm$ SD & NLL $\pm$ SD & Energy AUC $\pm$ SD \\
        \midrule
        Text only & 62.8 $\pm$ 0.7 & 0.842 $\pm$ 0.011 & 0.681 $\pm$ 0.010 \\
        Text plus $z$ & 68.9 $\pm$ 0.6 & 0.768 $\pm$ 0.010 & 0.732 $\pm$ 0.009 \\
        Text plus shuffled $z$ & 60.9 $\pm$ 0.8 & 0.874 $\pm$ 0.013 & 0.641 $\pm$ 0.012 \\
        Text plus zero $z$ & 63.4 $\pm$ 0.7 & 0.831 $\pm$ 0.012 & 0.687 $\pm$ 0.011 \\
        \bottomrule
    \end{tabular}
\end{table}

\begin{table}[t]
    \centering
    \caption{Parameter-matched controls for the lifted coordinate on Qwen3.5-27B. All variants use the same candidate budget, training budget, and three-seed protocol. The final column reports the paired aggregate gap between full LSPO and each control.}
    \label{tab:coordinate_controls}
    \small
    \setlength{\tabcolsep}{3.5pt}
    \begin{tabular}{lccc}
        \toprule
        Variant & Agg. $\pm$ SD & Net gain & LSPO $-$ variant (95\% CI) \\
        \midrule
        Full LSPO & 71.8 $\pm$ 0.15 & 4.7 & -- \\
        \midrule
        \multicolumn{4}{l}{\textit{Reference baseline (\Cref{tab:ablation_27b})}} \\
        Current no-lift baseline & 70.9 $\pm$ 0.18 & 3.2 & +0.9 [0.5, 1.3] \\
        \midrule
        \multicolumn{4}{l}{\textit{Parameter-matched coordinate controls}} \\
        Same-parameter no-lift MLP & 70.8 $\pm$ 0.18 & 3.0 & +1.0 [0.6, 1.4] \\
        Random $z$, matched dim. & 70.3 $\pm$ 0.19 & 2.8 & +1.5 [1.1, 1.9] \\
        Learned nuisance coordinate & 70.6 $\pm$ 0.18 & 2.9 & +1.2 [0.8, 1.6] \\
        Frozen / random projection & 70.4 $\pm$ 0.18 & 2.9 & +1.4 [1.0, 1.8] \\
        \bottomrule
    \end{tabular}%
\end{table}

\begin{table}[t]
    \centering
    \caption{Component ablation on the Qwen3.5-35B-A3B MoE backbone. The dense 27B model remains the primary attribution setting; this table tests whether the component pattern transfers to the MoE model.}
    \label{tab:moe_ablation}
    \small
    \begin{tabular}{lccccc}
        \toprule
        Variant & Agg. & Gain & Latency (s) & Tokens & GPU hours \\
        \midrule
        LSPO full & 72.6 & 4.4 & 8.30 & 1097 & 2410 \\
        Without lifting & 71.6 & 2.9 & 8.88 & 1141 & 2265 \\
        Without progress reward & 70.8 & 1.3 & 9.38 & 1185 & 2193 \\
        Without internalization & 71.1 & 2.1 & 9.13 & 1163 & 2073 \\
        Outcome RL & 71.5 & 2.7 & 9.33 & 1200 & 1940 \\
        DPO & 68.8 & 1.5 & 8.08 & 1048 & 990 \\
        \bottomrule
    \end{tabular}
\end{table}

\subsection{Training Configuration}

The following configuration specifies the primary training setting and the sensitivity ranges used in the component analyses.

The primary 27B LSPO run follows a fixed configuration. Reported means use seeds 42, 1234, and 3407. The objective coefficients are $\beta = 0.02$, $\lambda_{\mathrm{edit}} = 0.10$, and $\lambda_{\mathrm{step}} = 0.01$. The environment uses a maximum path length of 6, an auxiliary dimension of $8$, an imitation weight of 0.3, temperature 0.6, and top-$p = 0.95$. Optimization uses the Muon optimizer across all trainable parameters \citep{jordan2024muon,liu2025muon}. The configuration sets a learning rate of $2\times 10^{-5}$, momentum $0.95$, Nesterov updates, five Newton--Schulz steps, weight decay 0.1, and gradient clipping at 1.0. The schedule applies $500$ warmup steps and cosine decay to $10^{-7}$ over a $15{,}000$ step global horizon.

The benchmark suite and evaluation metrics remain fixed. Deployment evaluation measures pass@1 with direct generation or execution success. Supporting diagnostics include pass@16, transition rates, energy, latency, and output length.

\paragraph{Reported Variation.}
For the main experiments and training ablations, $\pm$ denotes the sample standard deviation across the three independent training seeds. We first average each seed's results over the complete evaluation set, then compute SD across those three means, using divisor $3-1$. The selection-budget audit in \Cref{tab:selection_budget_audit} instead fixes one checkpoint and reports bootstrap SD over evaluation inputs.

\subsection{Hyperparameter Selection}

The 27B hyperparameter search considers $\beta \in \{0.005, 0.01, 0.02, 0.05, 0.1\}$, $\lambda_{\mathrm{edit}} \in \{0.0, 0.05, 0.1, 0.2, 0.4\}$, and $\lambda_{\mathrm{step}} \in \{0.0, 0.005, 0.01, 0.02, 0.05\}$. Selection by aggregate validation accuracy under the three-seed protocol, with ties broken by minimized transition harm, yields $(0.02, 0.10, 0.01)$. These ranges describe model selection; the reported dimension ablations and joint candidate/horizon sweep provide the corresponding empirical sensitivity results.

Baseline models are instantiated under fixed matched settings. Standard preference optimization methods use minimal latency budgets; methods that operate on sampled paths consume longer outputs and higher latency budgets. LSPO narrows this efficiency gap by embedding machinery for local transitions into the parameters through answer internalization.

\subsection{Reward Decomposition}
\label{sec:appendix_progress_reward_theory}

The shaped path functional supplies the empirical ablations with a formal counterpart. \Cref{prop:reward_decomposition}, established in \Cref{sec:method}, shows that the progress reward telescopes into final energy improvement minus accrued edit and step costs, and \Cref{cor:path_preference} explains the preference for short, localized repairs over destructive global rewrites.

\paragraph{Telescoping identity.}
Here $T$ counts executed Revise/Switch moves. A terminal Stop is a zero-reward self-transition and leaves the identities unchanged. Substituting the move reward from \cref{eq:progress_reward} gives
\begin{equation}
\begin{aligned}
    \sum_{t=0}^{T-1} r_t
    &=
    \sum_{t=0}^{T-1}\big(\tilde E(s_t; q) - \tilde E(s_{t+1}; q)\big)
    - \lambda_{\mathrm{edit}} \sum_{t=0}^{T-1} d(x_{t+1}, x_t)
    - \lambda_{\mathrm{step}} T \\
    &=
    \tilde E(s_0; q) - \tilde E(s_T; q)
    - \lambda_{\mathrm{edit}} \sum_{t=0}^{T-1} d(x_{t+1}, x_t)
    - \lambda_{\mathrm{step}} T.
\end{aligned}
\end{equation}

\begin{corollary}[Preference for economical paths]
\label{cor:path_preference}
Let $\tau$ and $\tau'$ be two trajectories for the same prompt $q$ with identical initial and terminal states, satisfying $s_0 = s_0'$ and $s_{T_\tau} = s_{T_{\tau'}}'$. If
\begin{equation}
    \sum_{t=0}^{T_\tau-1} d(x_{t+1}, x_t) \le \sum_{t=0}^{T_{\tau'}-1} d(x_{t+1}', x_t')
    \qquad \text{and} \qquad
    T_\tau \le T_{\tau'},
\end{equation}
then the cumulative shaped reward satisfies $R(\tau; q) \ge R(\tau'; q)$. The inequality is strict whenever at least one inequality is strict and the corresponding coefficient $\lambda_{\mathrm{edit}}$ or $\lambda_{\mathrm{step}}$ is strictly positive.
\end{corollary}

\paragraph{Path economy argument.}
Under identical endpoints ($s_0 = s_0'$ and $s_{T_\tau} = s_{T_{\tau'}}'$), the net energy term $\tilde E(s_0; q) - \tilde E(s_T; q)$ is invariant across paths. Under this telescoping relation, $R(\tau; q) - R(\tau'; q)$ reduces entirely to differences in cumulative edit and step penalties, favoring trajectories that achieve the same descent with fewer tokens altered and fewer transitions taken.

The difference in energy aligns formally with potential reward shaping \citep{ng1999policy}. The edit and step penalties restrict the preferred policy class by penalizing unnecessarily long or destructive correction paths, favoring economical paths with net energy improvement.

\subsection{Noise Robustness}
\label{sec:appendix_stability_theory}

The ablation with noisy energy motivates a sensitivity analysis. Because the LSPO path reward relies on local energy differences, any bounded state perturbation results in a bounded endpoint error.

\begin{proposition}[Bounded energy noise perturbs cumulative reward only through endpoint error]
\label{prop:noisy_energy_robustness}
Let
\begin{equation}
    \hat E(s; q) = \tilde E(s; q) + \delta(s; q)
\end{equation}
with $|\delta(s; q)| \le \varepsilon$ for every visited state. Let $R(\tau; q)$ denote the cumulative shaped reward computed from $\tilde E$, and let $\hat R(\tau; q)$ denote the cumulative shaped reward computed from $\hat E$ using the same edit and step penalties. For any trajectory $\tau = (s_0, a_0, \dots, s_T)$,
\begin{equation}
    \hat R(\tau; q) - R(\tau; q)
    =
    \delta(s_0; q) - \delta(s_T; q),
    \qquad \text{and} \qquad
    \left|\hat R(\tau; q) - R(\tau; q)\right| \le 2\varepsilon.
\end{equation}
\end{proposition}

\paragraph{Endpoint perturbation.}
By \cref{prop:reward_decomposition},
\begin{equation}
    R(\tau; q)
    =
    \tilde E(s_0; q) - \tilde E(s_T; q)
    - \lambda_{\mathrm{edit}} \sum_{t=0}^{T-1} d(x_{t+1}, x_t)
    - \lambda_{\mathrm{step}} T
\end{equation}
and
\begin{equation}
    \hat R(\tau; q)
    =
    \hat E(s_0; q) - \hat E(s_T; q)
    - \lambda_{\mathrm{edit}} \sum_{t=0}^{T-1} d(x_{t+1}, x_t)
    - \lambda_{\mathrm{step}} T.
\end{equation}
Subtracting the two expressions cancels the penalty terms and leaves
\begin{equation}
    \hat R(\tau; q) - R(\tau; q)
    =
    \delta(s_0; q) - \delta(s_T; q).
\end{equation}
The bound uses $|\delta| \le \varepsilon$ at the two endpoints.

\begin{corollary}[Trajectory ranking is preserved under a sufficient terminal margin]
\label{cor:noisy_energy_ranking}
Let $\tau$ and $\tau'$ be two trajectories for the same prompt that start from the same initial state $s_0$. Under the conditions of \cref{prop:noisy_energy_robustness}, if
\begin{equation}
    R(\tau; q) - R(\tau'; q) > 2\varepsilon,
\end{equation}
the perturbed ranking satisfies $\hat R(\tau; q) > \hat R(\tau'; q)$.
\end{corollary}

\paragraph{Ranking margin.}
Because the two trajectories share the same initial state,
\begin{equation}
    \big(\hat R(\tau; q) - \hat R(\tau'; q)\big)
    -
    \big(R(\tau; q) - R(\tau'; q)\big)
    =
    -\delta(s_T; q) + \delta(s_T'; q).
\end{equation}
Its absolute value is at most $2\varepsilon$, giving
\begin{equation}
    \hat R(\tau; q) - \hat R(\tau'; q)
    \ge
    R(\tau; q) - R(\tau'; q) - 2\varepsilon
    >
    0,
\end{equation}
so the perturbed ranking is retained.

This bound gives a sufficient stability condition for a fixed candidate set under uniformly bounded state-wise corruption. The Gaussian-noise ablation provides a separate empirical sensitivity test. Applying the deterministic bound to any noise realization requires that all evaluated perturbations satisfy the stated limit.

\section{Internalization and Stopping Diagnostics}
\label{app:internalization}

This appendix evaluates the alignment between training rollouts and direct policy execution, the step-level monotonicity of energy descent, and the formal mechanics governing stopping conditions and answer internalization.

\begin{figure}[t]
    \centering
    \includegraphics[width=\linewidth]{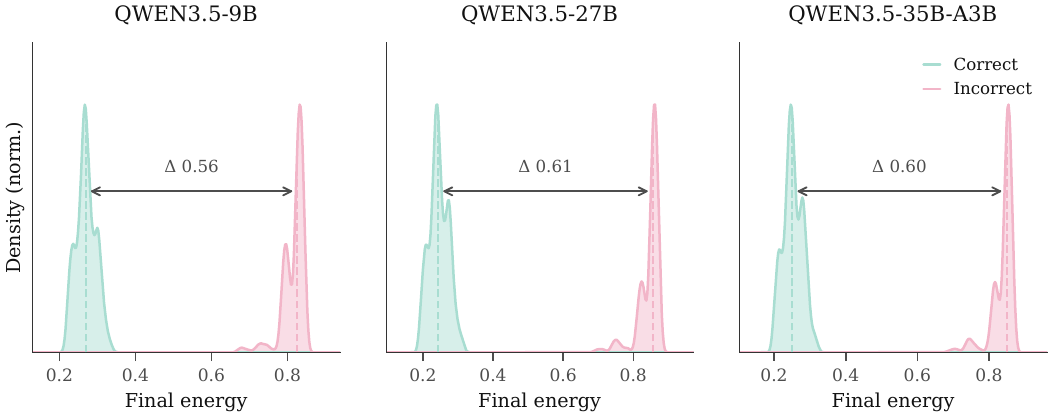}
\caption{Energy discriminates correct from incorrect endpoints across scales. Correct outputs consistently concentrate in regions with lower energy.}
    \label{fig:appendix_energy_correctness}
\end{figure}

\begin{figure}[t]
    \centering
    \includegraphics[width=\linewidth]{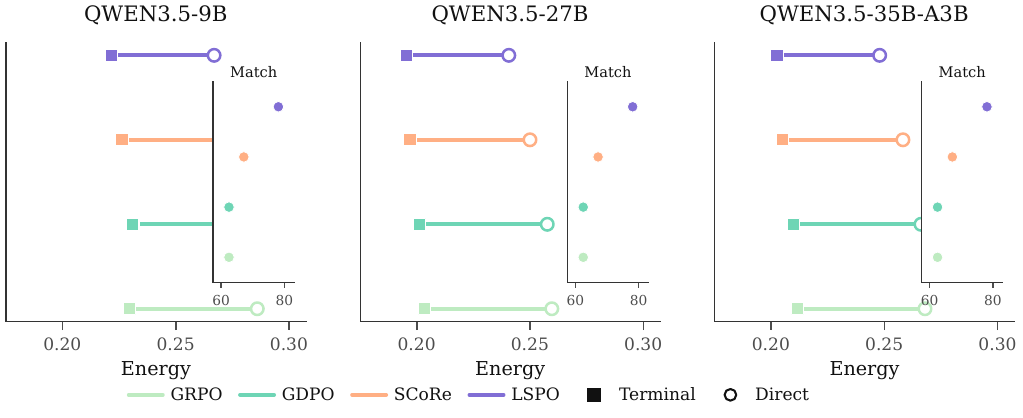}
    \caption{Training endpoints align with direct outputs after LSPO. The divergence between terminal states selected during alignment and direct generations is minimal under the LSPO objective.}
    \label{fig:appendix_rollout_direct_alignment}
\end{figure}

\begin{figure*}[t]
    \centering
    \includegraphics[width=\textwidth]{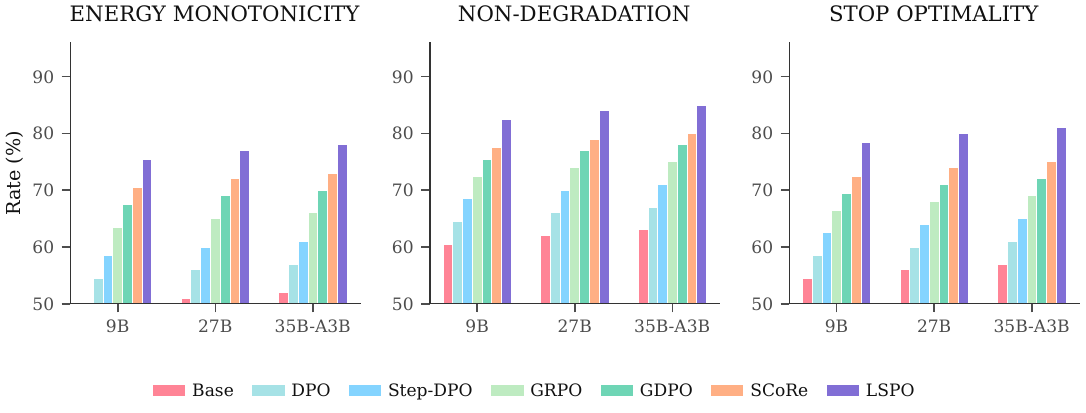}
\caption{Step diagnostics across scales. LSPO preserves the strongest monotonicity in energy descent, correct local structure, and stop optimality across the evaluated backbones.}
    \label{fig:appendix_monotonicity_diagnostics}
\end{figure*}

\begin{figure*}[t]
    \centering
    \includegraphics[width=\textwidth]{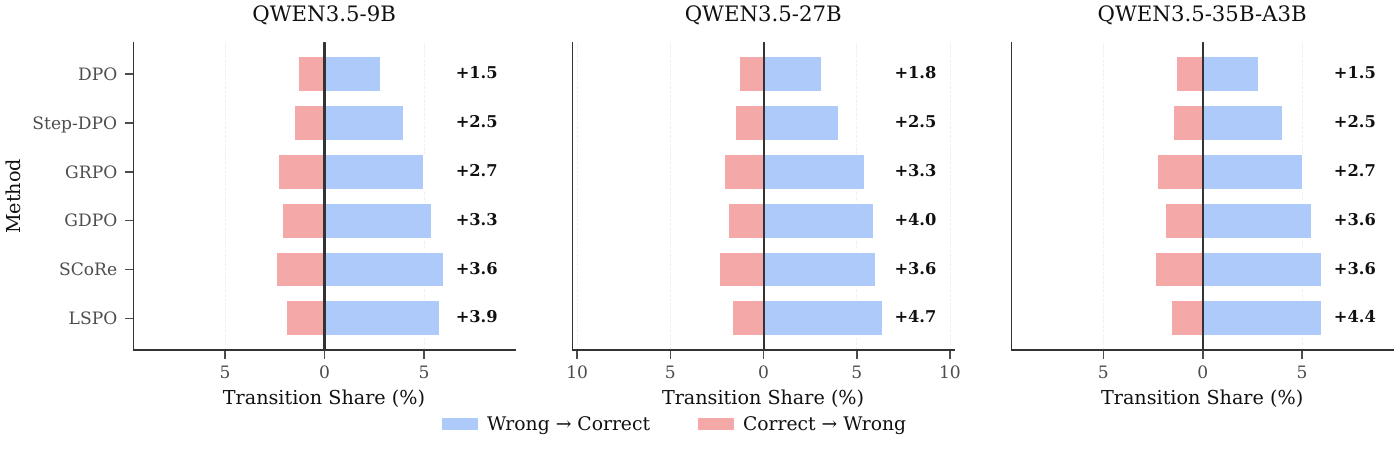}
\caption{Transition gains across scales. LSPO achieves the largest surplus of conversions from incorrect to correct on every backbone while keeping conversions from correct to incorrect modest.}
    \label{fig:appendix_transition_breakdown_all_models}
\end{figure*}

\Cref{fig:appendix_energy_correctness} evaluates energy as an ordering signal. Across all three model scales, lower final energy is associated with higher correctness, though slopes and separation margins vary by capacity. We therefore compare absolute energy values within each backbone, consistent with prior work on model calibration and uncertainty \citep{kadavath2022language, yin2023large, guo2017calibration}.

\Cref{fig:appendix_rollout_direct_alignment} separates training exploration from deployed policy outputs. LSPO shows the smallest energy gap between selected training answers and direct outputs, the empirical signature of answer internalization. The analysis measures whether correction traces shift the policy distribution, complementing prior studies of self-correction \citep{kumar2024training, ma2025s2r, xiong2025self}.

\Cref{fig:appendix_monotonicity_diagnostics} resolves the mechanism at the step level. The energy monotonicity rate measures how often a move decreases state energy; the rate of preserving correct states measures whether already correct states are retained; and the stop optimality rate evaluates whether generation ends near the empirical minimum along the visited path. The ordering of these diagnostics follows the geometric constraints encouraged by LSPO.

\Cref{fig:appendix_transition_breakdown_all_models} extends the transition diagnostic beyond the primary 27B setting. The ordering persists at 9B and 35B-A3B. LSPO increases constructive conversions from incorrect to correct without the large conversions from correct to incorrect observed in search baselines with high cost.

\subsection{Answer and Span Transition Metrics}
\label{sec:transition_metrics}

For each method, answer-level transitions compare the before and after correctness labels of the same $N$ answer pairs. Let $n_{W\to C}$ and $n_{C\to W}$ count repairs (wrong to correct) and regressions (correct to wrong), respectively. Both reported rates use the full paired sample as denominator,
\begin{equation}
G_{\mathrm{ans}}=100\,n_{W\to C}/N,\qquad
H_{\mathrm{ans}}=100\,n_{C\to W}/N,\qquad
\mathrm{NetGain}=G_{\mathrm{ans}}-H_{\mathrm{ans}}.
\end{equation}
Thus the 6.4\% repair and 1.7\% regression frequencies for LSPO yield a 4.7-point net gain. Correct-to-correct (C$\to$C) and incorrect-to-incorrect (W$\to$W) pairs contribute zero to this difference.

The span audit in \Cref{tab:span_transitions} uses before/after validity labels, denoted $V$ and $I$, for paired spans. Each method has its own initial valid fraction $p_V$. Preservation and damage condition on initially valid spans, while repair conditions on initially invalid spans,
\begin{equation}
\begin{aligned}
P_{\mathrm{span}}&=\Pr(V_{\mathrm{after}}\mid V_{\mathrm{before}}),&
D_{\mathrm{span}}&=1-P_{\mathrm{span}},\\
R_{\mathrm{span}}&=\Pr(V_{\mathrm{after}}\mid I_{\mathrm{before}}),&
\mathrm{NetSpanGain}&=100[(1-p_V)R_{\mathrm{span}}-p_VD_{\mathrm{span}}].
\end{aligned}
\end{equation}
Probabilities in this equation lie in $[0,1]$; net gain is in percentage points. Full LSPO preserves 95.8\% of initially valid spans and repairs 29.5\% of initially invalid spans, yielding a 16.0-point increase in the valid-span fraction. The matched no-lift control yields 9.9 points. These results support preservation and repair of the spans assessed by the validity labels; answer-level and span-level gains describe distinct units of analysis.

\begin{table}[t]
    \centering
    \caption{Span validity transitions. Initial validity, preservation, repair, and damage are percentages; net gain is in percentage points. Preservation and damage condition on initially valid spans; repair conditions on initially invalid spans. Net gain weights these rates by each method's initial valid fraction.}
    \label{tab:span_transitions}
    \small
    \setlength{\tabcolsep}{4pt}
    \begin{tabular}{lrrrrr}
        \toprule
        Method & Initial $p_V$ & $V\to V$ & $I\to V$ & $V\to I$ & Net gain \\
        \midrule
        Full LSPO & 40.1 & \textbf{95.8} & \textbf{29.5} & \textbf{4.2} & \textbf{+16.0} \\
        Same-parameter no-lift MLP & 39.7 & 92.1 & 21.6 & 7.9 & +9.9 \\
        Without progress reward & 39.8 & 89.8 & 18.7 & 10.2 & +7.2 \\
        Terminal-only reward & 39.1 & 90.3 & 16.9 & 9.7 & +6.5 \\
        GRPO & 39.4 & 90.9 & 18.6 & 9.1 & +7.7 \\
        \bottomrule
    \end{tabular}
\end{table}

\subsection{Stopping Condition}
\label{sec:appendix_stopping_internalization_theory}

The stop optimality diagnostic has a formal counterpart. Continuing from a state cannot improve cumulative reward when every future energy decrease is exhausted by the associated edit and step costs.

\begin{proposition}[A sufficient stopping condition]
\label{prop:stopping_condition}
Fix an input $q$ and a move index $t$, counting Revise/Switch transitions. Suppose that for every admissible continuation
\begin{equation}
    s_t, s_{t+1}, \dots, s_T
\end{equation}
starting from $s_t$, each future transition indexed by $u \in \{t,\dots,T-1\}$ satisfies
\begin{equation}
    \tilde E(s_u; q) - \tilde E(s_{u+1}; q)
    \le
    \lambda_{\mathrm{edit}} d(x_{u+1}, x_u) + \lambda_{\mathrm{step}}.
\end{equation}
Every such continuation has nonpositive additional cumulative reward,
\begin{equation}
    \sum_{u=t}^{T-1} r_u \le 0.
\end{equation}
For the unregularized shaped return, stopping with zero reward weakly dominates these continuations. The action-KL penalty in \cref{eq:rl_objective} can change the regularized policy preference.
\end{proposition}

\paragraph{Stopping argument.}
The assumed inequality implies
\begin{equation}
    \tilde E(s_u; q) - \tilde E(s_{u+1}; q)
    - \lambda_{\mathrm{edit}} d(x_{u+1}, x_u)
    - \lambda_{\mathrm{step}}
    \le 0
\end{equation}
for every $u \ge t$, i.e., $r_u \le 0$. Summing over $u=t,\dots,T-1$ gives nonpositive cumulative reward.

\subsection{Internalization Projection}
\label{sec:appendix_internalization_theory}

The alignment figures motivate a formal interpretation of the answer internalization objective. Training selects an answer distribution with lower energy, and the deployed generator is then optimized toward it as closely as model capacity permits. For each input $q$, hold fixed the sampled-target distribution $\mu_q$ of selected outputs $x_T^*(q)$ induced by the selection procedure during training.

\begin{proposition}[Answer internalization as forward KL projection]
\label{prop:si_projection}
For each input $q$,
\begin{equation}
    \mathbb{E}_{x \sim \mu_q}\!\left[\log p_\theta(x \mid q)\right]
    =
    - H(\mu_q)
    -
    D_{\mathrm{KL}}\!\big(\mu_q \,\|\, p_\theta(\cdot \mid q)\big),
\end{equation}
so maximizing $J_{\mathrm{SI}}(\theta)$ is equivalent, up to the parameter-independent constant $\mathbb{E}_{q}[H(\mu_q)]$, to minimizing the expected forward KL to the selected-answer distributions.
\end{proposition}

\paragraph{Projection identity.}
Substituting $\mathbb{E}_{x \sim \mu_q}[\log \mu_q(x)] = -H(\mu_q)$ into the definition of forward relative entropy yields the identity; averaging over $q$ and dropping the constant term gives the equivalence.

\begin{proposition}[Bounded task scores are controlled by forward KL on the selected-answer distribution]
\label{prop:rollout_direct_gap}
Fix an input $q$ and let $c_q(x) \in [0,1]$ denote any correctness or task success score for output $x$. Define
\begin{equation}
    \Delta_q(\theta)
    =
    \mathbb{E}_{x \sim \mu_q}[c_q(x)]
    -
    \mathbb{E}_{x \sim p_\theta(\cdot \mid q)}[c_q(x)].
\end{equation}
The gap for each input obeys
\begin{equation}
    \left|\Delta_q(\theta)\right|
    \le
    \mathrm{TV}\!\big(\mu_q, p_\theta(\cdot \mid q)\big)
    \le
    \sqrt{
        \frac{1}{2}
        D_{\mathrm{KL}}\!\big(\mu_q \,\|\, p_\theta(\cdot \mid q)\big)
    }.
\end{equation}
Averaging over inputs gives
\begin{equation}
    \mathbb{E}_{q}\!\left[\left|\Delta_q(\theta)\right|\right]
    \le
    \sqrt{
        \frac{1}{2}
        \mathbb{E}_{q}\!\left[
            D_{\mathrm{KL}}\!\big(\mu_q \,\|\, p_\theta(\cdot \mid q)\big)
        \right]
    }.
\end{equation}
\end{proposition}

\paragraph{Total variation bound.}
For any function $f \in [0,1]$,
\begin{equation}
    \left|
        \mathbb{E}_{\mu_q}[f(x)] - \mathbb{E}_{p_\theta(\cdot \mid q)}[f(x)]
    \right|
    \le
    \mathrm{TV}\!\big(\mu_q, p_\theta(\cdot \mid q)\big).
\end{equation}
Applying this to $f=c_q$ gives the first bound. The second bound is Pinsker's inequality,
\begin{equation}
    \mathrm{TV}\!\big(\mu_q, p_\theta(\cdot \mid q)\big)
    \le
    \sqrt{
        \frac{1}{2}
        D_{\mathrm{KL}}\!\big(\mu_q \,\|\, p_\theta(\cdot \mid q)\big)
    }.
\end{equation}
Finally, averaging over $q$ and using Jensen's inequality for the concave square root function yields
\begin{equation}
    \mathbb{E}_{q}\!\left[\left|\Delta_q(\theta)\right|\right]
    \le
    \mathbb{E}_{q}\!\left[
        \sqrt{
            \frac{1}{2}
            D_{\mathrm{KL}}\!\big(\mu_q \,\|\, p_\theta(\cdot \mid q)\big)
        }
    \right]
    \le
    \sqrt{
        \frac{1}{2}
        \mathbb{E}_{q}\!\left[
            D_{\mathrm{KL}}\!\big(\mu_q \,\|\, p_\theta(\cdot \mid q)\big)
        \right]
    }.
\end{equation}

For each fixed target distribution, the bound links the alignment diagnostic in \Cref{fig:appendix_rollout_direct_alignment} to the training objective, showing that as the forward KL term decreases, the expected gap in bounded task success scores is correspondingly controlled. It conditions on a fixed $\mu_q$ and is informative only where the generator can approach that distribution; thus, we use it as a consistency interpretation of the alignment figures rather than as an accuracy guarantee.

\subsection{Comparison with Test-Time Reflection}

Test-time reflection systems retain critique and revision within the inference loop \citep{jiang2026reflexicoder}. LSPO generates and selects training traces before deployment, then serves answers with the direct generator. The alignment diagnostics above measure how much of the resulting improvement persists in that generator, while the deployment comparison quantifies the cost of retaining deliberation at inference.

\section{Deployment Complexity and Held-Out Energy Ordering}
\label{app:energy_complexity}

Deployment cost, energy ordering on tasks held out from training, and compute accounting summarize LSPO's operational profile. The diagnostics characterize energy as a local ordering signal and quantify the inference cost that remains after correction traces are absorbed into the model.

\begin{figure}[t]
    \centering
    \includegraphics[width=\linewidth]{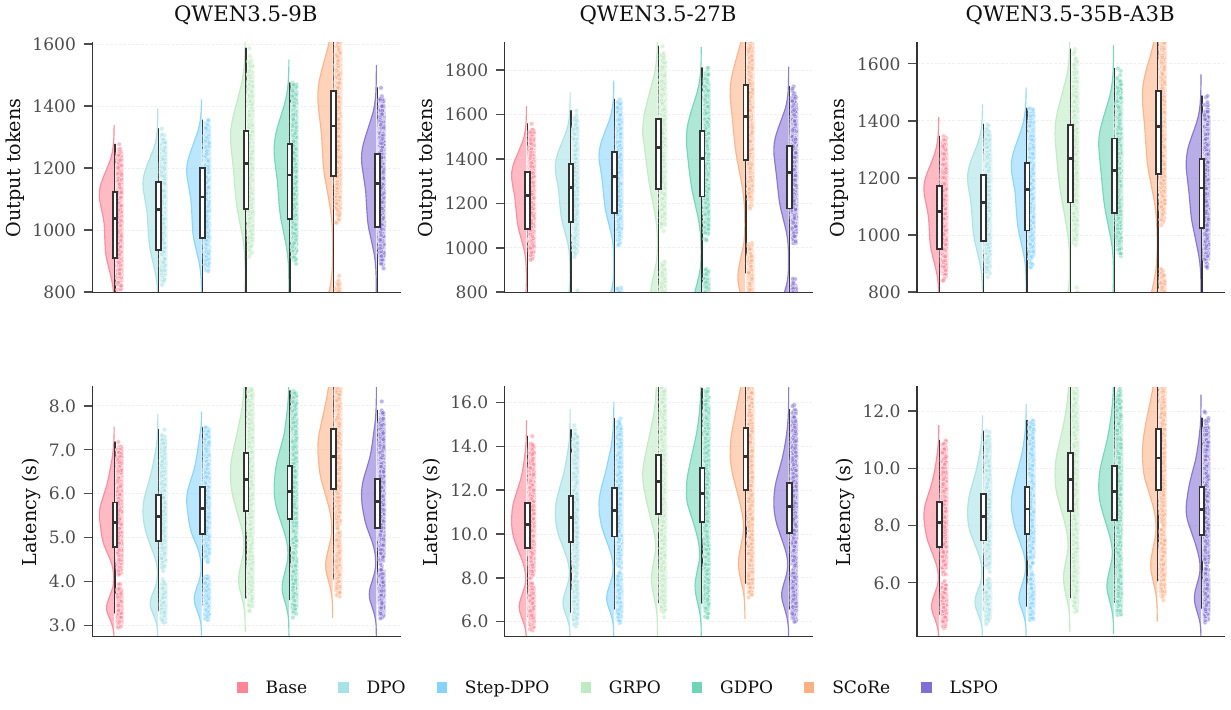}
\caption{Deployment cost distributions. LSPO stays within the latency and length ranges established by preference optimization, remaining below baselines that use extensive search.}
    \label{fig:appendix_cost_distributions}
\end{figure}

\Cref{fig:appendix_cost_distributions} quantifies the deployment cost of the direct policy. Systems that use extensive search can incur large latency and verbosity penalties \citep{chen2021evaluating, guo2025deepseek}. LSPO produces slightly longer generations than standard preference methods such as DPO, yet remains below the token and latency costs of GRPO and SCoRe. The method stays within a practical deployment range while preserving the accuracy gains reported in the main text.

The next analyses study the relationship between state energy and empirical correctness. When lower scores correlate with correct outputs, the scalar can serve as both a shaping signal during training and a diagnostic of candidate quality. Prior work on calibration fragility \citep{niculescu2005predicting, guo2017calibration} and semantic uncertainty \citep{kuhn2023semantic, farquhar2024detecting, kossen2024semantic, aichberger2024improving} motivates interpreting energy as a structural indicator of local state ordering.

\subsection{Energy Ordering on Held-Out Tasks}
\label{sec:appendix_energy_calibration}

The empirical correlation between low energy and correctness reported in \Cref{fig:appendix_energy_correctness} motivates a conditional calibration analysis. Under a shared prompt-wise monotone calibration map, an almost-sure energy ordering under a coupling implies an ordering of expected correctness.

All ordering metrics evaluate the utility score $u(s,q)=-\tilde E(s;q)$, where higher $u$ indicates lower energy and a preferred candidate under the learned ordering. The AUC between energy and correctness measures the probability that a correct candidate receives a higher $u$ than an incorrect candidate \citep{fawcett2006introduction}. Spearman and Kendall compute rank correlations between $u$ and the binary correctness label over the evaluated candidate set, with Kendall using a rank statistic corrected for ties \citep{spearman1961proof,kendall1938new}. Pairwise preference accuracy calculates the fraction of differently labeled candidate pairs within each prompt where the correct candidate has lower energy, aligning with the standard pairwise ranking logic of preference models \citep{bradley1952rank}. The Bins metric partitions candidates into five equal-frequency energy bins and reports the length of the contiguous monotone prefix, measuring how many of the five bins preserve the expected decrease in empirical correctness as energy rises. ECE partitions candidates into bins with equal numbers via a frozen monotonic calibration map and reports the expected absolute difference between the predicted and empirical correctness probabilities \citep{naeini2015obtaining,guo2017calibration}.

\begin{figure*}[t]
    \centering
    \includegraphics[width=.98\textwidth]{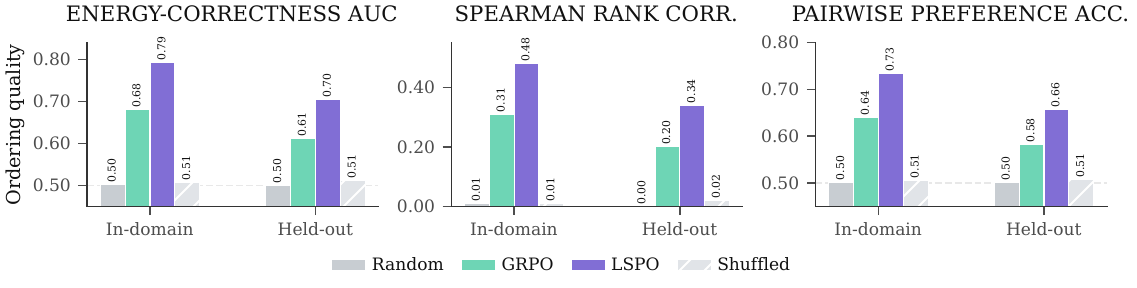}
\caption{Frozen energy ordering under distribution shift. On tasks held out from fitting (GSM8K, ARC-Challenge, BBH, HumanEval, and MBPP), the LSPO energy head outperforms outcome only, shuffled, and random controls in area under the curve, Spearman correlation, and pairwise accuracy.}
    \label{fig:heldout_energy_ordering}
\end{figure*}

\begin{table}[t]
    \centering
    \caption{Frozen energy ordering under distribution shift. LSPO yields the strongest AUC, rank correlation, and pairwise preference accuracy on both the fitted task families and tasks held out from fitting.}
    \label{tab:heldout_energy_ordering}
    \small
    \begin{tabular}{lcccccc}
        \toprule
        Score & AUC & Spearman & Kendall & Pairwise & Bins & ECE \\
        \midrule
        \multicolumn{7}{l}{\textit{Task families used for fitting include math, code, logic, and science}} \\
        Random energy & 0.502 & 0.01 & 0.01 & 0.502 & 2 & 0.241 \\
        Outcome only score & 0.681 & 0.31 & 0.22 & 0.641 & 4 & 0.132 \\
        LSPO energy & 0.792 & 0.48 & 0.35 & 0.734 & 5 & 0.071 \\
        Shuffled LSPO energy & 0.508 & 0.01 & 0.01 & 0.506 & 2 & 0.235 \\
        \midrule
        \multicolumn{7}{l}{\textit{Task families held out from fitting include GSM8K, ARC-C, BBH, HumanEval, and MBPP}} \\
        Random energy & 0.501 & 0.00 & 0.00 & 0.501 & 2 & 0.247 \\
        Outcome only score & 0.612 & 0.20 & 0.14 & 0.583 & 3 & 0.170 \\
        LSPO energy & 0.704 & 0.34 & 0.24 & 0.658 & 4 & 0.112 \\
        Shuffled LSPO energy & 0.511 & 0.02 & 0.01 & 0.507 & 2 & 0.238 \\
        \bottomrule
    \end{tabular}
\end{table}

\paragraph{Data Separation and Locked Evaluation.}
\Cref{tab:training_corpus} identifies the public training corpora. The main evaluation suites are MATH-500, AIME25, GPQA, LiveCodeBench, ZebraLogic, and MMLU-Pro. GSM8K \citep{cobbe2021training}, ARC-Challenge \citep{clark2018think}, BBH \citep{suzgun2023challenging}, HumanEval \citep{chen2021evaluating}, and MBPP \citep{austin2021program} provide additional benchmark-level holdouts for transfer and frozen energy-ordering audits. Both groups follow the same exclusion rules.

Domain overlap is permitted. Evaluation examples and near-duplicates are excluded from policy training, energy fitting, pairwise ranking construction, training-target selection, verifier/checker threshold tuning, hyperparameter search, checkpoint selection, selection of training and audit budgets, and construction of prompts, rubrics, or synthetic references. Use of other examples from the same benchmark requires explicit deduplication against the evaluation set.

Evaluation questions enter only the locked final evaluation. Generation and candidate selection use the fixed model and scoring branch without gold answers, correctness labels, test-time verifiers, or judge feedback. External checkers and judges score the resulting outputs for reporting. Decoding settings and the training budget $K$ and audit budgets $K_{\mathrm{sel}}$ are fixed before evaluation; test results serve only to report performance across those budgets.

As reported in \Cref{fig:heldout_energy_ordering} and \Cref{tab:heldout_energy_ordering}, the domain shift reduces the LSPO energy AUC from 0.792 to 0.704 and pairwise preference accuracy from 0.734 to 0.658. Both metrics remain above the outcome only and shuffled controls, supporting energy as a local ranking signal beyond the fitted task families.

\subsection{Direct Generation on Tasks Outside the Training Distribution}

The evaluation on benchmarks held out from fitting also tests the deployed policy itself. The suites in \Cref{tab:ood_direct_generation} follow the exclusion rules above, including exclusion from training-target selection. LSPO improves over Base on all five tasks and reaches an average of 74.9 on held-out tasks, above SCoRe at 74.3, without a regression on any task relative to Base. The held-out policy results complement the ordering audit: the frozen energy head retains local signal under shift, and the compiled policy transfers the corresponding improvement gains to direct generation.

\begin{table}[t]
    \centering
    \caption{Direct generation on benchmark suites held out from policy training, energy fitting, and endpoint selection. LSPO improves over Base on every benchmark.}
    \label{tab:ood_direct_generation}
    \small
    \begin{tabular}{lrrrrrr}
        \toprule
        Method & GSM8K & ARC-C & BBH & HumanEval & MBPP & Avg. $\pm$ SD \\
        \midrule
        Base & 88.4 & 84.7 & 66.2 & 55.5 & 62.4 & 71.4 $\pm$ 0.22 \\
        DPO & 89.0 & 85.3 & 67.1 & 56.4 & 63.0 & 72.2 $\pm$ 0.20 \\
        GRPO & 90.6 & 86.7 & 69.0 & 58.6 & 64.8 & 73.9 $\pm$ 0.22 \\
        SCoRe & 91.1 & 87.1 & 69.3 & 59.1 & 65.0 & 74.3 $\pm$ 0.24 \\
        \midrule
        LSPO & 91.5 & 87.4 & 70.4 & 60.1 & 65.2 & 74.9 $\pm$ 0.20 \\
        \bottomrule
    \end{tabular}
\end{table}

\subsection{Joint Candidate and Horizon Sensitivity}

Candidate quantity and transition horizon affect the local candidate space explored during alignment. Here $K$ is the maximum total number of candidate states retained per input $q$, including the initial direct-generation state $s_0$. \Cref{tab:candidate_horizon} varies $K$ and $T_{\max}$ together under the same training and evaluation protocol, measuring sensitivity to their joint budget. The gain persists at every tested budget; aggregate accuracy reaches 71.8 at the primary configuration $(T_{\max},K)=(6,7)$ and remains at that value for $(8,9)$. The larger budget increases realized path length and reduces the ratio of unique candidates.

\begin{table}[t]
    \centering
    \caption{Joint candidate and horizon sensitivity on Qwen3.5-27B. $K$ counts the maximum number of retained states per input, including the initial direct state $s_0$. Values are means $\pm$ sample SD across three independent training seeds. Aggregate accuracy reaches 71.8 at $(T_{\max},K)=(6,7)$ and remains at that value for $(8,9)$.}
    \label{tab:candidate_horizon}
    \small
    \begin{tabular}{cccccc}
        \toprule
        $T_{\max}$ & $K$ & Agg. $\pm$ SD & Gain $\pm$ SD & Path length & Unique ratio \\
        \midrule
        2 & 3 & 70.9 $\pm$ 0.18 & 3.2 $\pm$ 0.30 & 1.6 & 0.82 \\
        4 & 5 & 71.5 $\pm$ 0.16 & 4.2 $\pm$ 0.26 & 2.7 & 0.76 \\
        6 & 7 & 71.8 $\pm$ 0.15 & 4.7 $\pm$ 0.23 & 3.4 & 0.71 \\
        8 & 9 & 71.8 $\pm$ 0.16 & 4.8 $\pm$ 0.24 & 3.7 & 0.66 \\
        \bottomrule
    \end{tabular}
\end{table}

\begin{table}[t]
    \centering
    \caption{Policy transfer pilot on subjective tasks. LSPO maintains positive transfer over Base across both benchmarks without requiring test-time rubrics, reference answers, or verifier loops.}
    \label{tab:subjective_transfer}
    \small
    \begin{tabular}{lccccc}
        \toprule
        Method & WritingBench L\&A & $\Delta$ Base & MT-Bench Roleplay & $\Delta$ Base & avg. \\
        \midrule
        Base & 7.42 $\pm$ 0.08 & 0.00 & 7.61 $\pm$ 0.11 & 0.00 & 7.52 \\
        DPO & 7.50 $\pm$ 0.07 & +0.08 & 7.68 $\pm$ 0.10 & +0.07 & 7.59 \\
        GRPO & 7.36 $\pm$ 0.09 & -0.06 & 7.53 $\pm$ 0.12 & -0.08 & 7.45 \\
        SCoRe & 7.32 $\pm$ 0.09 & -0.10 & 7.49 $\pm$ 0.12 & -0.12 & 7.41 \\
        DRO-adapted & 7.53 $\pm$ 0.08 & +0.11 & 7.70 $\pm$ 0.10 & +0.09 & 7.62 \\
        \midrule
        LSPO & 7.48 $\pm$ 0.08 & +0.06 & 7.66 $\pm$ 0.10 & +0.05 & 7.57 \\
        \bottomrule
    \end{tabular}
\end{table}

\begin{proposition}[Accuracy improvement under monotone energy calibration]
\label{prop:energy_calibration}
Fix a prompt $q$ and two endpoint distributions $\mu_q,\nu_q$. Assume a common measurable nonincreasing map $\phi_q$ calibrates correctness under each distribution, so that for $\kappa\in\{\mu_q,\nu_q\}$
\begin{equation}
    \mathbb{P}_{S\sim\kappa}\big(\mathrm{correct}=1 \mid \tilde E(S; q)=e, q\big) = \phi_q(e).
\end{equation}
Suppose there exists a coupling $(S_\mu, S_\nu)$ with marginals $\mu_q$ and $\nu_q$ such that
\begin{equation}
    \tilde E(S_\mu; q) \le \tilde E(S_\nu; q)
\end{equation}
holds with probability one. Then the coupled endpoints satisfy
\begin{equation}
    \mathbb{E}_{S_\mu \sim \mu_q}\!\left[\phi_q\big(\tilde E(S_\mu; q)\big)\right]
    \ge
    \mathbb{E}_{S_\nu \sim \nu_q}\!\left[\phi_q\big(\tilde E(S_\nu; q)\big)\right].
\end{equation}
\end{proposition}

\paragraph{Coupling argument.}
Under the assumed coupling and the monotonicity of $\phi_q$, the pointwise ordering $\phi_q(\tilde E(S_\mu; q)) \ge \phi_q(\tilde E(S_\nu; q))$ holds almost surely. Taking expectations directly preserves the inequality.

The common calibration map and coupled ordering are sufficient conditions for the correctness comparison. An ordering of mean energy alone is insufficient. The empirical ordering audits assess local ranking utility and leave these distribution-level assumptions conditional.

\subsection{Energy Semantics and Local Ordering}

The energy $\tilde E$ is a learned scalar head trained for state ranking. Unlike global uncertainty measures such as entropy or Fisher information \citep{shannon1948mathematical, renyi1961measures, fisher1925theory, rao1945information, cramer1946contribution}, it serves as a local diagnostic of state quality, where lower energy identifies a preferred candidate within the immediate vicinity.

Recent evaluations of factuality in short form text show that empirical correctness depends on measurement protocols specific to each domain \citep{wei2024measuring}. The scalar carries evidence here because it tracks constructive revisions within the transition geometry induced by LSPO. We pair energy analysis with transition diagnostics to evaluate both state ordering and answer changes.

\subsection{Open Form Transfer}

The open form pilot measures policy transfer on subjective tasks. We use the Literature and Arts subset of WritingBench \citep{wu2026writingbench} and the Roleplay subset of MT-Bench \citep{zheng2023judging}. Each method generates one response per turn without revision or candidate selection, and the same judge model and protocol score all outputs. We also evaluate DRO-adapted, based on Direct Reasoning Optimization \citep{xu2025direct}. DRO optimizes unverifiable reasoning via token-level self-certainty rewards (R3), query-specific rubric gating, and variance-based filtering over GRPO. Because the original framework targets reference-based tasks like ParaRev, RaR-Medicine, ContractNLI, and FinQA, we adapt it to open form evaluation by supplying synthesized references and task rubrics.

As shown in \Cref{tab:subjective_transfer}, training paradigms diverge sharply on subjective tasks. Standard RL with GRPO and self-correction with SCoRe degrade below Base to combined averages of 7.45 and 7.41, as optimizing exclusively for verifiable correctness weakens open-ended expressiveness. Preference tuning with DPO maintains conversational quality at 7.59. DRO-adapted reaches the highest combined average of 7.62 by leveraging rubric gating and reference guidance. Meanwhile, LSPO achieves positive transfer across both suites at a 7.57 combined average, clearly exceeding GRPO and SCoRe while remaining close to DPO. Crucially, while DRO-adapted depends on synthesized references and rubrics, LSPO operates in a clean direct generation regime with zero test-time references, rubrics, or verifier loops.

For a more local audit, we score seven candidates from the same prompt with the frozen energy head and use judge scores only as evaluation labels. We compute correlations within each input and pairwise ordering, then average those quantities across prompts. \Cref{tab:open_form_energy_audit} shows moderate alignment, strongest for instruction fit, coherence, and character consistency. \Cref{tab:open_form_failure_slices} identifies prompts requiring high creativity and playful improvisation, where the learned ordering is less reliable. Energy supports local assessment of constrained style and consistency; literary quality and expressive richness require additional signals.

\begin{table}[t]
    \centering
    \caption{Frozen energy ordering in open form transfer. Energy is evaluated as an ordering of seven candidates from each input; judge scores are used only for evaluation.}
    \label{tab:open_form_energy_audit}
    \small
    \begin{tabular}{lcccc}
        \toprule
        Subset & Spearman $\rho$ & Kendall $\tau$ & Pairwise AUC & $\Delta$ \\
        \midrule
        WritingBench Literature and Arts & 0.34 $\pm$ 0.04 & 0.24 $\pm$ 0.03 & 0.675 $\pm$ 0.018 & +0.42 \\
        MT-Bench Roleplay & 0.29 $\pm$ 0.06 & 0.20 $\pm$ 0.04 & 0.642 $\pm$ 0.026 & +0.35 \\
        \bottomrule
    \end{tabular}
\end{table}

\begin{table}[t]
    \centering
    \caption{Alignment dimensions and failure slices for the open form audit. The frozen energy ordering is most reliable on constrained dimensions, while open-ended creativity and expressive richness remain its main failure modes.}
    \label{tab:open_form_failure_slices}
    \small
    \resizebox{\textwidth}{!}{
    \begin{tabular}{lccc}
        \toprule
        Subset & Strongest alignment & Weakest alignment & Failure slice \\
        \midrule
        WritingBench Literature and Arts & Instruction fit, coherence & Creativity, originality & Prompts requiring high creativity \\
        MT-Bench Roleplay & Character consistency, instruction fit & Expressive richness & Playful improvisation \\
        \bottomrule
    \end{tabular}
    }
\end{table}

\begin{table}[t]
    \centering
    \caption{Compute and deployment accounting for Qwen3.5-27B. Training uses $8\times$ NVIDIA A100 80GB GPUs. The LSPO cost of 2983.2 GPU-hours is the mean per-run cost across three seeds.}
    \label{tab:compute_disclosure}
    \small
    \resizebox{\textwidth}{!}{%
    \begin{tabular}{lcccccc}
        \toprule
        Method & Peak mem (GB) & Elapsed time (h) & GPU hrs & Pilot GPU hours & Latency (s) & Output tokens \\
        \midrule
        DPO & 57.7 & 152.2 & 1217.6 & 146.1 & 10.42 & 1196 \\
        Step-DPO & 57.7 & 203.1 & 1624.8 & 195.0 & 10.73 & 1242 \\
        GRPO & 57.7 & 298.3 & 2386.4 & 286.4 & 12.04 & 1370 \\
        GDPO & 57.7 & 335.6 & 2684.8 & 322.2 & 11.54 & 1324 \\
        SCoRe & 57.7 & 358.2 & 2865.7 & 343.9 & 13.12 & 1501 \\
        \midrule
        LSPO & 61.9 & 372.9 & 2983.2 & 358.0 & 10.93 & 1265 \\
        \bottomrule
    \end{tabular}
    }
\end{table}

\begin{table}[t]
    \centering
    \caption{Energy ordering under increasing selection budgets with one fixed Full LSPO checkpoint. $K_{\mathrm{sel}}$ counts additional candidates beyond the direct output; selection uses all $K_{\mathrm{sel}}+1$ candidates. Energy and open-form scores are means $\pm$ bootstrap SD over evaluation inputs. Mean energy averages one selected state per input. Energy drop and MATH-500 change are relative to direct generation; open-form scores are absolute.}
    \label{tab:selection_budget_audit}
    \small
    \begin{tabular}{crrrrr}
        \toprule
        $K_{\mathrm{sel}}$ & Mean $\tilde E$ & Energy drop & \makecell{MATH-500\\$\Delta$ EM (pp)} & \makecell{WritingBench\\L\&A} & \makecell{MT-Bench\\Roleplay} \\
        \midrule
        0  & $0.398\pm0.0041$ & 0.0\%  & 0.0  & $7.47\pm0.077$ & $7.65\pm0.104$ \\
        1  & $0.322\pm0.0044$ & 19.1\% & +0.2 & $7.50\pm0.083$ & $7.68\pm0.109$ \\
        2  & $0.268\pm0.0047$ & 32.7\% & +0.6 & $7.53\pm0.091$ & $7.71\pm0.115$ \\
        4  & $0.224\pm0.0049$ & 43.7\% & +0.4 & $7.53\pm0.087$ & $7.71\pm0.121$ \\
        8  & $0.195\pm0.0052$ & 51.0\% & +0.2 & $7.50\pm0.096$ & $7.68\pm0.128$ \\
        16 & $0.171\pm0.0057$ & 57.0\% & -0.4 & $7.45\pm0.111$ & $7.62\pm0.141$ \\
        \bottomrule
    \end{tabular}
\end{table}

\subsection{Compute Disclosure}

Training uses $8\times$ NVIDIA A100 80GB GPUs. The reported 2983.2 GPU-hours for LSPO is the mean cost of one training run across seeds 42, 1234, and 3407. Engineering evaluation requires concrete resource accounting alongside complexity analysis. \Cref{tab:compute_disclosure} documents the primary 27B hardware footprint and deployment cost in one view, including peak memory, elapsed time, total GPU hours, pilot budgets, latency, and output length.

The resource table fixes the balance between training and deployment. LSPO spends more during training than SCoRe, adding 117.5 GPU hours on 27B. At deployment it avoids explicit critique, revision, or search, reducing average latency from 13.12 to 10.93\,s and output length from 1501 to 1265 tokens.

\subsection{Energy Ordering under Increasing Selection Budgets}
\label{sec:selection_budget_audit}

We audit how the frozen energy ordering behaves as the candidate pool expands around a fixed trained policy. The deployment results use direct generation ($K_{\mathrm{sel}}=0$); this diagnostic varies additional candidate generation while holding the checkpoint and scoring branch fixed. The evaluation-time budget $K_{\mathrm{sel}}$ counts additional candidates generated beyond the direct output, giving $K_{\mathrm{sel}}+1$ candidates in the selection pool. For positive budgets, we select the minimum-energy state from the pool, including the direct output. The joint $(T_{\max},K)$ sweep above concerns retained training states; this audit concerns additional generation after training.

For each evaluation input $q_i$, let $\mathcal C_i(K_{\mathrm{sel}})$ be its candidate pool and select $s_i^\star=\arg\min_{s\in\mathcal C_i(K_{\mathrm{sel}})}\tilde E(s;q_i)$. The energy column averages exactly one selected state per input over all $N$ audit inputs,
\begin{equation}
    \overline E(K_{\mathrm{sel}})=\frac{1}{N}\sum_{i=1}^{N}\tilde E(s_i^\star;q_i).
\end{equation}
All inputs contribute regardless of selected-answer correctness. Energy and open-form score entries report means $\pm$ bootstrap SD, obtained by resampling evaluation inputs with the checkpoint fixed. The SD therefore describes sampling variation over inputs.

\Cref{tab:selection_budget_audit} reports MATH-500 exact-match changes in percentage points relative to direct generation, alongside absolute WritingBench Literature and Arts and MT-Bench Roleplay scores. On the 500 MATH-500 questions, the changes correspond to $0,+1,+3,+2,+1,-2$ correct answers relative to direct generation for $K_{\mathrm{sel}}=0,1,2,4,8,16$, respectively. Energy reductions use the fixed scoring scale in this audit. Increasing the budget continues to lower selected energy, while external quality peaks at smaller budgets and declines at the largest budget tested.

From $K_{\mathrm{sel}}=8$ to $16$, mean selected energy falls from $0.195$ to $0.171$, a 12.3\% reduction, while MATH-500 exact match falls by 0.6 points and the two open-form scores fall by 0.05 and 0.06. These are observed changes in means. The continued decrease in selected energy alongside the decline in external-quality means at larger budgets identifies a boundary of the learned ordering under increased selection pressure. These observations qualify the use of energy as a local ranking signal over the evaluated candidate distributions.

\section{Qualitative Case Studies and Failure Modes}
\label{app:qualitative}

These cases translate the quantitative diagnostics into four repair regimes covering targeted mathematical correction, bounded code repair, preservation of correct local logic, and global backtracking.

\subsection{Repair Regimes}

The mathematics and logic examples exhibit the pattern measured by the transition metrics. Successful revisions change the flawed step while preserving surrounding correct structure. LSPO fits this regime, where the desired update is nearby and the energy signal can rank adjacent states.

The code examples highlight excessive editing as a distinct failure mode. RL over sampled paths can discover executable solutions at the cost of broad rewrites. The edit and step penalties in LSPO bias optimization toward economical patches that leave the surrounding logic intact, providing a case example of the tighter latency and output length distributions reported in \Cref{app:energy_complexity}.

\subsection{Global Search Failure Mode}

Case 4 marks the principal failure mode. LSPO repairs nearby slips within a flawed path, whereas problems requiring exhaustive exploration may need a procedure that rebuilds a global search tree at inference. Geometric shaping is most effective when the required improvement lies near the current state, consistent with prior work on self-correction \citep{madaan2023self, shinn2023reflexion, huang2023large, ma2025s2r}.

\vspace{1em}

\definecolor{pastelBlueBg}{RGB}{240,244,248}
\definecolor{pastelBlueFr}{RGB}{108,136,162}

\definecolor{pastelGreenBg}{RGB}{242,247,242}
\definecolor{pastelGreenFr}{RGB}{127,156,134}

\definecolor{pastelPurpleBg}{RGB}{245,242,248}
\definecolor{pastelPurpleFr}{RGB}{135,123,158}

\definecolor{pastelRoseBg}{RGB}{252,240,241}
\definecolor{pastelRoseFr}{RGB}{194,125,128}

\begin{tcolorbox}[
    colback=pastelBlueBg,
    colframe=pastelBlueFr,
    coltitle=white,
    title=\textbf{Case 1: Local Math Repair (MATH-500)},
    arc=2mm,
    boxrule=0.5mm,
    left=2mm, right=2mm, top=2mm, bottom=2mm
]
\textbf{Prompt Summary}\\
Algebraic simplification with a late arithmetic slip

\vspace{0.5em}
\textbf{Step-DPO Baseline}\\
Rewrites the derivation but preserves the same arithmetic mistake. \textit{Supervision over outputs rewards restatement without isolating the faulty update.}

\vspace{0.5em}
\textbf{LSPO}\\
Preserves the earlier derivation and repairs only the erroneous scalar update.

\textbf{\textit{Interpretation.}} \textit{A repair from incorrect to correct (W$\to$C) follows a local correction associated with lower energy.}
\end{tcolorbox}

\begin{tcolorbox}[
    colback=pastelGreenBg,
    colframe=pastelGreenFr,
    coltitle=white,
    title=\textbf{Case 2: Bounded Code Repair (LiveCodeBench)},
    arc=2mm,
    boxrule=0.5mm,
    left=2mm, right=2mm, top=2mm, bottom=2mm
]
\textbf{Prompt Summary}\\
Repair a boundary condition bug in a short Python function

\vspace{0.5em}
\textbf{GRPO Baseline}\\
Fixes the bug via a larger refactor that alters unrelated control flow. \textit{Sparse outcome rewards permit changes outside the failing boundary.}

\vspace{0.5em}
\textbf{LSPO}\\
Applies a short local guard and preserves the rest of the function.

\textbf{\textit{Interpretation.}} \textit{Edit and step costs penalize unnecessary rewrite breadth.}
\end{tcolorbox}

\begin{tcolorbox}[
    colback=pastelPurpleBg,
    colframe=pastelPurpleFr,
    coltitle=white,
    title=\textbf{Case 3: Local Logic Preservation (ZebraLogic)},
    arc=2mm,
    boxrule=0.5mm,
    left=2mm, right=2mm, top=1.5mm, bottom=1.5mm
]
\textbf{Prompt Summary}\\
Resolve a constraint puzzle with one correct partial assignment

\vspace{0.3em}
\textbf{DPO Baseline}\\
Rewrites the assignment and loses a previously consistent clue. \textit{Preference optimization over answers risks structural regressions from correct to incorrect.}

\vspace{0.3em}
\textbf{LSPO}\\
Keeps the valid partial assignment and flips only the conflicting slot.

\textbf{\textit{Interpretation.}} \textit{Transition shaping preserves already correct local structure.}
\end{tcolorbox}

\begin{tcolorbox}[
    colback=pastelRoseBg,
    colframe=pastelRoseFr,
    coltitle=white,
    title=\textbf{Case 4: Global Backtracking (AIME25)},
    arc=2mm,
    boxrule=0.5mm,
    left=2mm, right=2mm, top=1.5mm, bottom=1.5mm
]
\textbf{Prompt Summary}\\
Competition math problem requiring exhaustive branch-and-bound search

\vspace{0.3em}
\textbf{Test-Time Reflection Baseline}\\
Discards the initial derivation entirely and generates an alternative branching tree. \textit{Unconstrained inference loops can escape global failure states.}

\vspace{0.3em}
\textbf{LSPO}\\
Repairs a minor algebraic slip but remains trapped in the incorrect logical branch.

\textbf{\textit{Interpretation.}} \textit{Internalization through direct generation improves nearby repairs; global backtracking requires broader exploration.}
\end{tcolorbox}


\newpage
\section*{NeurIPS Paper Checklist}

\begin{enumerate}

\item {\bf Claims}
    \item[] Question: Do the main claims made in the abstract and introduction accurately reflect the paper's contributions and scope?
    \item[] Answer: \answerYes{}
    \item[] Justification: The abstract and introduction present cost-adjusted transition credit, lifted-state learning, and answer internalization, with the primary 27B accuracy and deployment-cost comparison. \Cref{sec:method} formalizes the objective, \Cref{sec:experiments} evaluates it across dense and mixture-of-experts Qwen3.5 backbones, and \Cref{sec:limitations} states its scope and limitations.

\item {\bf Limitations}
    \item[] Question: Does the paper discuss the limitations of the work performed by the authors?
    \item[] Answer: \answerYes{}
    \item[] Justification: \Cref{sec:limitations} discusses dependence on energy-ordering fidelity, the local-repair scope, upfront training cost, and the current restriction of the auxiliary coordinate to textual deliberation and mainly verifiable reasoning tasks.

\item {\bf Theory assumptions and proofs}
    \item[] Question: For each theoretical result, does the paper provide the full set of assumptions and a complete (and correct) proof?
    \item[] Answer: \answerYes{}
    \item[] Justification: The formal results are stated with explicit assumptions and proved in Appendices~\ref{app:theory_roadmap},~\ref{app:ablation_stability},~\ref{app:internalization}, and~\ref{app:energy_complexity}. The main-text equations (\Cref{eq:progress_reward,eq:rl_objective,eq:si_objective,eq:lspo_objective}) are accompanied by propositions and corollaries with complete derivations in those appendices.

\item {\bf Experimental result reproducibility}
    \item[] Question: Does the paper fully disclose all the information needed to reproduce the main experimental results of the paper to the extent that it affects the main claims and/or conclusions of the paper (regardless of whether the code and data are provided or not)?
    \item[] Answer: \answerYes{}
    \item[] Justification: Appendix~\ref{app:ablation_stability} provides LSPO pseudocode and the fixed seed set, objective coefficients, path-length limit, optimizer settings, and validation/checkpoint schedule. The public implementation and configuration are available at \url{https://github.com/Venomrko/lspo}.
    \item[] Guidelines:
    \begin{itemize}
        \item The answer \answerNA{} means that the paper does not include experiments.
        \item If the paper includes experiments, a \answerNo{} answer to this question will not be perceived well by the reviewers, because reproducibility is important regardless of whether the code and data are provided.
        \item If the contribution is a dataset and\slash or model, the authors should describe the steps taken to make their results reproducible or verifiable.
        \item Depending on the contribution, reproducibility can be accomplished in various ways. For example, if the contribution is a novel architecture, describing the architecture fully might suffice, or if the contribution is a specific model and empirical evaluation, it may be necessary to either make it possible for others to replicate the model with the same dataset, or provide access to the model. In general. releasing code and data is often one good way to accomplish this, but reproducibility can also be provided via detailed instructions for how to replicate the results, access to a hosted model (e.g., in the case of a large language model), releasing of a model checkpoint, or other means that are appropriate to the research performed.
        \item While NeurIPS does not require releasing code, the conference does require all submissions to provide some reasonable avenue for reproducibility, which may depend on the nature of the contribution. For example
        \begin{enumerate}
            \item If the contribution is primarily a new algorithm, the paper should make it clear how to reproduce that algorithm.
            \item If the contribution is primarily a new model architecture, the paper should describe the architecture clearly and fully.
            \item If the contribution is a new model (e.g., a large language model), then there should either be a way to access this model for reproducing the results or a way to reproduce the model (e.g., with an open-source dataset or instructions for how to construct the dataset).
            \item We recognize that reproducibility may be tricky in some cases, in which case authors are welcome to describe the particular way they provide for reproducibility. In the case of closed-source models, it may be that access to the model is limited in some way (e.g., to registered users), but it should be possible for other researchers to have some path to reproducing or verifying the results.
        \end{enumerate}
    \end{itemize}

\item {\bf Open access to data and code}
    \item[] Question: Does the paper provide open access to the data and code, with sufficient instructions to faithfully reproduce the main experimental results, as described in supplemental material?
    \item[] Answer: \answerYes{}
    \item[] Justification: The public repository at \url{https://github.com/Venomrko/lspo} provides the LSPO implementation, training and evaluation scripts, baseline and ablation scripts, dependencies, paper configuration, public training-data identifiers, and setup and execution commands. Appendix~\ref{app:ablation_stability} documents the fixed experimental protocol.

\item {\bf Experimental setting/details}
    \item[] Question: Does the paper specify all the training and test details (e.g., data splits, hyperparameters, how they were chosen, type of optimizer) necessary to understand the results?
    \item[] Answer: \answerYes{}
    \item[] Justification: \Cref{sec:experiments} gives the model and benchmark setup. Appendix~\ref{app:ablation_stability} specifies the Muon optimizer settings, learning rate, momentum, weight decay, path-length limit, internalization weight, hyperparameter sweep ranges, selected primary configuration, seed set, and validation/checkpoint cadence.

\item {\bf Experiment statistical significance}
    \item[] Question: Does the paper report error bars suitably and correctly defined or other appropriate information about the statistical significance of the experiments?
    \item[] Answer: \answerYes{}
    \item[] Justification: The main and appendix tables report sample standard deviation across independent training seeds 42, 1234, and 3407, with paired confidence intervals for selected controls. Appendix~\ref{app:ablation_stability} defines how the seed-level standard deviation is calculated; the fixed-checkpoint selection audit reports bootstrap standard deviation over evaluation inputs.

\item {\bf Experiments compute resources}
    \item[] Question: For each experiment, does the paper provide sufficient information on the computer resources (type of compute workers, memory, time of execution) needed to reproduce the experiments?
    \item[] Answer: \answerYes{}
    \item[] Justification: Appendix~\ref{app:energy_complexity} provides a compute-disclosure table for the primary 27B comparisons, covering the GPU type, peak memory, elapsed time, GPU-hours, and pilot GPU-hours. The main-text tables and cost plots report deployment latency and output length.

\item {\bf Code of ethics}
    \item[] Question: Does the research conducted in the paper conform, in every respect, with the NeurIPS Code of Ethics \url{https://neurips.cc/public/EthicsGuidelines}?
    \item[] Answer: \answerYes{}
    \item[] Justification: To the best of our knowledge, the work conforms with the NeurIPS Code of Ethics. It uses existing model families and publicly available benchmarks and does not involve human subjects.

\item {\bf Broader impacts}
    \item[] Question: Does the paper discuss both potential positive societal impacts and negative societal impacts of the work performed?
    \item[] Answer: \answerNo{}
    \item[] Justification: \Cref{sec:limitations} discusses deployment efficiency and technical failure modes, but the paper does not explicitly assess both positive and negative societal impacts beyond those technical considerations.

\item {\bf Safeguards}
    \item[] Question: Does the paper describe safeguards that have been put in place for responsible release of data or models that have a high risk for misuse (e.g., pre-trained language models, image generators, or scraped datasets)?
    \item[] Answer: \answerNA{}
    \item[] Justification: The current release provides research code, but no new model weights, dataset, or public inference endpoint with a high risk for misuse.

\item {\bf Licenses for existing assets}
    \item[] Question: Are the creators or original owners of assets (e.g., code, data, models), used in the paper, properly credited and are the license and terms of use explicitly mentioned and properly respected?
    \item[] Answer: \answerYes{}
    \item[] Justification: The paper credits the model families, benchmarks, datasets, and prior methods used. The public repository identifies its training-data sources, notes that model weights and datasets retain their respective licenses, and releases the original LSPO code under the MIT license. Third-party model weights and datasets are not redistributed.

\item {\bf New assets}
    \item[] Question: Are new assets introduced in the paper well documented and is the documentation provided alongside the assets?
    \item[] Answer: \answerYes{}
    \item[] Justification: The new LSPO software is available at \url{https://github.com/Venomrko/lspo} with an MIT license, dependency list, paper configuration, data-source identifiers, hardware requirements, and training and evaluation instructions. No new dataset or model checkpoint is released.

\item {\bf Crowdsourcing and research with human subjects}
    \item[] Question: For crowdsourcing experiments and research with human subjects, does the paper include the full text of instructions given to participants and screenshots, if applicable, as well as details about compensation (if any)?
    \item[] Answer: \answerNA{}
    \item[] Justification: The work does not involve crowdsourcing or research with human subjects.

\item {\bf Institutional review board (IRB) approvals or equivalent for research with human subjects}
    \item[] Question: Does the paper describe potential risks incurred by study participants, whether such risks were disclosed to the subjects, and whether Institutional Review Board (IRB) approvals (or an equivalent approval/review based on the requirements of your country or institution) were obtained?
    \item[] Answer: \answerNA{}
    \item[] Justification: The work does not involve crowdsourcing or research with human subjects, so IRB approval is not applicable.

\item {\bf Declaration of LLM usage}
    \item[] Question: Does the paper describe the usage of LLMs if it is an important, original, or non-standard component of the core methods in this research? Note that if the LLM is used only for writing, editing, or formatting purposes and does \emph{not} impact the core methodology, scientific rigor, or originality of the research, declaration is not required.
    \item[] Answer: \answerYes{}
    \item[] Justification: The core method post-trains and evaluates large language models. The model family, post-training setup, and deployment configuration are described in \Cref{sec:method,sec:experiments}.

\end{enumerate}

\end{document}